\documentclass[journal]{IEEEtran}
\ifCLASSINFOpdf
\else
\fi
\usepackage{microtype}
\usepackage{graphicx}
\usepackage{subfigure}
\usepackage{booktabs} 
\usepackage[table]{xcolor}

\usepackage[ruled]{algorithm2e}
\usepackage{algorithmic}
\usepackage{amsmath}
\usepackage{multicol}
\usepackage{makecell}
\usepackage[outline]{contour}

\usepackage{tabu}
\usepackage{multirow}

\usepackage{hyperref}
\usepackage{arydshln}

\usepackage{amsmath, amssymb, amsthm}
\usepackage{physics}

\newcommand{\D}{\mathbf{D}}

\newcommand{\bigO}{\mathcal{O}}
\newcommand{\R}{\mathbb{R}}
\newcommand{\Set}{\mathcal{S}}
\newcommand{\W}{\mathbf{W}}
\newcommand{\U}{\mathbf{U}}
\newcommand{\V}{\mathbf{V}}

\newcommand{\ai}{\mathbf{a}}

\newcommand{\bias}{\mathbf{b}}

\newcommand{\loss}{\ell} 
\newcommand{\p}{\mathbf{p}}
\newcommand{\q}{\mathbf{q}}
\newcommand{\m}{\mathbf{m}}
\newcommand{\x}{\mathbf{x}}

\newcommand{\z}{\mathbf{z}}

\newcommand{\up}{\mathbf{u}}

\newcommand{\low}{\mathbf{l}}

\DeclareMathOperator*{\argmin}{arg\,min} 

\usepackage{mathtools}
\DeclarePairedDelimiter\ceil{\lceil}{\rceil}

\usepackage[font=small,labelfont=bf]{caption}

\newtheorem{theory}{Theorem}
\newtheorem{lemma}{Lemma}

\contourlength{0.2pt}
\contournumber{20}
\newcommand{\change}[1]{{\color{black} #1}}
\newcommand{\minorchange}[1]{{\color{black} #1}}
\newcommand{\one}[1]{\underline{\textbf{#1}}}
\newcommand{\two}[1]{\textbf{#1}}

\newcommand{\eqdef}{\overset{\mathrm{def}}{=\joinrel=}}
\newcolumntype{P}[1]{>{\centering\arraybackslash}p{#1}}

\begin{document}
%
\title{Training Provably Robust Models \\ by Polyhedral Envelope Regularization}
%
%
%

\author{Chen~Liu, Student Member, IEEE, Mathieu~Salzmann, Member, IEEE and Sabine~S\"usstrunk, Fellow, IEEE
\thanks{C. Liu, M. Salzmann, S. S\"usstrunk are with the School of Computer and Communication Sciences, École polytechnique fédérale de Lausanne (EPFL), Lausanne, Switzerland.}}

\maketitle

\begin{abstract}

Training certifiable neural networks enables us to obtain models with robustness guarantees against adversarial attacks.
In this work, we introduce a framework to obtain a provable adversarial-free region in the neighborhood of the input data by a polyhedral envelope, which yields more fine-grained certified robustness than existing methods.
\change{
We further introduce polyhedral envelope regularization (PER) to encourage larger adversarial-free regions and thus improve the provable robustness of the models.
We demonstrate the flexibility and effectiveness of our framework on standard benchmarks; it applies to networks of different architectures and with general activation functions.
Compared with state of the art, PER has negligible computational overhead; it achieves better robustness guarantees and accuracy on the clean data in various settings.
}

\end{abstract}


\section{Introduction}

Despite their great success in many applications, modern deep learning models are vulnerable to adversarial attacks: small but well-designed perturbations can make the state-of-the-art models predict wrong labels with very high confidence \cite{goodfellow2014explaining, moosavi2017universal, szegedy2013intriguing}.
The existence of such adversarial examples indicates unsatisfactory properties of the deep learning models' decision boundary \cite{he2018decision}, and poses a threat to the reliability of safety-critical machine learning systems.

As a consequence, studying the robustness of deep learning has attracted growing attention, from the perspective of both attack and defense strategies.
Popular attack algorithms, such as the \textit{Fast Gradient Sign Method} (FGSM) \cite{goodfellow2014explaining}, the CW attack \cite{carlini2017towards} and the \textit{Projected Gradient Descent} (PGD) \cite{madry2017towards}, typically exploit the gradient of the loss w.r.t. the input to generate adversarial examples.
\change{Recently, the state-of-the-art success rates have been attained with adaptive methods, such as \textit{Auto Attack}\cite{croce2020reliable}.}
All these methods assume that the attackers have access to the model parameters and thus belong to the ``white-box attacks''.
On the contrary, ``black-box attacks'' tackle the cases where the attackers have limited access to the model, \change{such as limited access to the output logits \cite{andriushchenko2019square}), hard-label predictions (\cite{cheng2018queryefficient} and task settings \cite{dong2018boosting}}.

To counteract such attacks, robust learning aims to learn a model which optimizes the worst-case loss over the allowable perturbations.
Formally, given a model $f_\theta$ parameterized by $\theta$, the loss function $\loss$ and a dataset $\mathcal{D}$, robust learning solves the following min-max problem:

\begin{equation}
\begin{aligned}
\min_{\theta} \mathbb{E}_{(\x, y) \sim \mathcal{D}} \max_{\x' \in \Set_\epsilon(\x)} \loss(f_\theta(\x'), y)
\end{aligned} \label{eq:defadv}
\end{equation}

Here, $\Set_\epsilon(\x)$ denotes the adversarial budget, which is the allowable perturbed input of the clean input $\x$.
To solve (\ref{eq:defadv}), many defense algorithms have been proposed \cite{buckman2018thermometer, dhillon2018stochastic,ma2018characterizing,pang2020rethinking,pang2019improving,samangouei2018defense,xiao2020enhancing}.
However, it was shown by~\cite{athalye2018obfuscated,croce2020reliable,tramer2020adaptive} that most of them depend on obfuscated gradients for perceived robustness.
In other words, these methods train models to fool gradient-based attacks but do not achieve true robustness.
As a consequence, they become ineffective when subjected to stronger attacks.
The remaining effective defense is adversarial training \cite{madry2017towards} and its extensions \cite{alayrac2019labels,carmon2019unlabeled,gowal2020uncovering,hendrycks2019using,mao2019metric,sinha2019harnessing,wu2020adversarial,zhang2019you}, which augments the training data with adversarial examples.
Nevertheless, while adversarial training yields good empirical performance under adaptive attacks, it still provides no \textit{guarantees} of a model's robustness.

\change{In this work, we focus on constructing certifiers to find \textit{certified regions} of the input neighborhood where the model is \textit{guaranteed} to give the correct prediction, and on using such certifiers to train a model to be \textit{provably} robust against adversarial attacks.}
To obtain such robustness guarantee, there are two categories of methods: complete certifiers and incomplete certifiers.
Complete certifiers can either guarantee the absence of an adversary or find an adversarial example given an adversarial budget.
They are typically built on either \textit{Satisfiability Modulo Theories} (SMT) \cite{katz2017reluplex} or \textit{Mixed Integer Programming} (MIP) \cite{tjeng2017evaluating, xiao2018training}.
The major disadvantages of complete certifiers are their super-polynomial complexity and applicability to only piecewise linear activation functions, such as ReLU.
By contrast, incomplete certifiers are faster, more widely applicable but more conservative in terms of certified regions because they rely on approximations.
In this context, techniques such as linear approximation \cite{balunovic2020adversarial, kolter2017provable, weng2018towards, wong2018scaling, zhang2018efficient}, symbolic interval analysis \cite{wang2018efficient}, abstract transformers \cite{gehr2018ai2,singh2018fast,singh2019abstract} and semidefinite programming \cite{raghunathan2018certified, raghunathan2018semidefinite} have been exploited to offer better certified robustness.
In addition, recent works use randomized smoothing \cite{cohen2019certified, salman2019provably} to construct probabilistic certifiers, which provides robustness guarantees with high probability by Monte Carlo sampling.
Some of these methods enable training provably robust models \cite{cohen2019certified, kolter2017provable, raghunathan2018certified, salman2019provably, wong2018scaling} by optimizing the model parameters so as to maximize the area of the certified regions.

While effective, all the above-mentioned certification methods, \change{except for randomized smoothing, which gives probabilistic guarantees,} only provide binary results given a \textit{fixed} adversarial budget in their vanilla version.
That is, if a data point is certified, it is guaranteed to be robust in the entire given adversarial budget; otherwise no guaranteed adversary-free region is estimated.
To overcome this and search for the \textit{optimal size of the adversarial budget} that can be certified,~\cite{kolter2017provable, weng2018towards, zhang2018efficient} use either Newton's method or binary search.
By contrast,~\cite{croce2018provable} takes advantage of the geometric property of ReLU networks and gives more fine-grained robustness guarantees.
Based on the piecewise linear nature of the ReLU function, any input is located in a polytope where the network can be considered a linear function.
Based on geometry, robustness guarantees can thus be calculated using the input data's distance to the polytope boundary and the decision boundary constraints.
Unfortunately, in practice, the resulting certified bounds are trivial because such polytopes are very small even for robust models.
Nevertheless,~\cite{croce2018provable} introduces a regularization scheme based on these bounds, models trained using this regularizer are provably robust by other certifiers.

\change{
In this paper, we construct a stronger certifier, as well as a regularization scheme to train provably robust models.
Instead of relying on the linear regions of the ReLU networks, we estimate a linear bound on the model's output given a predefined adversarial budget.
Then, the condition to guarantee robustness inside this budget is also linear and forms a polyhedral envelope of the model's decision boundary.
The intersection of the polyhedral envelope and the predefined adversarial budget is then guaranteed to be adversary-free.
In contrast to~\cite{croce2018provable}, our method can be based on any model linearization method and is thus applicable to general network architectures and activation functions.
To train provably robust neural network models, we further introduce a hinge-loss-like regularization term to encourage larger certified bounds.
Furthermore, we boost the performance of our  method with adversarial training.
We also use a stochastic robust approximation \cite{wang2018mixtrain} to accelerate our method and reduce its memory consumption.

Based on the geometry of the decision boundary, our proposed certification method significantly improves the one in~\cite{croce2018provable} and yields a more accurate estimation of the decision boundary.
Furthermore, it is more generally applicable to different activation functions.
In contrast to Fast-Lin \cite{weng2018towards} and CROWN \cite{zhang2018efficient}, our certification method can prove that a subset of the adversarial budget is adversary-free.
We show that such partial credit can accelerate the search for the optimal size of the adversarial budget.
On the training side, in contrast to KW \cite{kolter2017provable, wong2018scaling}, which, as pointed out by~\cite{zhang2019towards}, over-regularizes the model, our proposed method achieves better certified robustness without sacrificing too much clean accuracy.
In the remainder of the paper, we refer to our certification method as \textit{Polyhedral Envelope Certifier (PEC)} and to our regularization scheme as \textit{Polyhedral Envelope Regularizer (PER)}.
}

\section{Preliminaries} \label{sec:preliminary}

\subsection{Notation and Terminology}

For simplicity, we discuss our approach using a standard $N$-layer fully-connected network.
\minorchange{We will discuss how this formulation can be extended to other architectures in Section~\ref{sec:model_linearization}.}
A fully-connected network parameterized by $\{\W^{(i)}, \bias^{(i)}\}_{i = 1}^{N - 1}$ can be expressed as the following equations:
\begin{equation}
\begin{aligned}
\z^{(i + 1)} &= \W^{(i)}\widehat{\z}^{(i)} + \bias^{(i)} & i = 1, 2, ..., N - 1 \\
\widehat{\z}^{(i)} &= \sigma(\z^{(i)}) & i = 2, 3, ..., N - 1
\end{aligned} \label{eq:model}
\end{equation}
where $\z^{(i)}$ and $\widehat{\z}^{(i)}$ are the pre- \& post-activations of the $i$-th layer, respectively, and $\widehat{\z}^{(1)} \eqdef \x$ is the input of the network.
An $l_p$ norm-based adversarial budget $\Set^{(p)}_\epsilon(\x)$ is defined as the set $\{\x' | \|\x' - \x\|_p \leq \epsilon\}$.
$\x'$, $\z'^{(i)}$ and $\widehat{\z}'^{(i)}$ represent the adversarial input and the corresponding pre- \& post-activations.
For layer $i$ having $n_i$ neurons, we have $\W^{(i)} \in \R^{n_{i + 1} \times n_i}$ and $\bias^{(i)} \in \R^{n_{i + 1}}$.
We use $K \eqdef n_{N}$ to represent the output dimension.

Throughout this paper, underlines and bars are used to represent lower and upper bounds of the corresponding variables, respectively, i.e., $\underline{\z}^{(i)} \leq \z'^{(i)} \leq \bar{\z}^{(i)}$.
A ``$+$'' or ``$-$'' subscript indicates the positive or negative elements of a tensor, with all other elements replacing with $0$.
We use $[K]$ as the abbreviation for the set $\{1, 2, ..., K\}$.

\subsection{Model Linearization} \label{sec:model_linearization}

Given an adversarial budget $\Set^{(p)}_\epsilon(\x)$, we study the linear bound of the output logits $\z'^{(N)}$, given by

\begin{equation}
\begin{aligned}
\U^{(N)}\x' + \p^{(N)} \leq \z'^{(N)} \leq \V^{(N)}\x' + \q^{(N)}\;.
\end{aligned} \label{eq:linearize}
\end{equation}

The linear coefficients introduced above can be calculated by iteratively estimating the bounds of intermediate layers and linearizing the activation functions.
In Appendix \ref{subsec:app_linearize_activation}, we discuss this for several activation functions, including ReLU, sigmoid and tanh.
Note that our method differs from~\cite{zhang2018efficient} as we need the analytical form of the linear coefficients for training.
For example~\cite{zhang2018efficient} uses some numerical methods such as binary search, while our method does not.
\change{
The bounding algorithm trades off computational complexity and bound tightness.
In this work, we study two such algorithms.
One, which we call \textit{CROWN-based bounds}, is based on Fast-Lin / CROWN \cite{weng2018towards, zhang2018efficient}.
It yields tighter bounds but has higher computational complexity.
The other, which we call \textit{IBP-inspired bounds}, is inspired by the \textit{Interval Bound Propagation} (IBP) \cite{gowal2018effectiveness}.
It is faster but leads to looser bounds.
The details of both algorithms are provided in Appendices~\ref{subsec:app_fastlin_crown} and~\ref{subsec:app_ibp}, respectively.
We discuss the complexity of both algorithms in detail in Section~\ref{sec:discussion}.
}

\minorchange{
Although the formulation above is based on the fully-connected network, it can be straightforwardly extend to any network whose corresponding computational graph can be represented by a \textit{Directed Acyclic Graph}.
All the factors in our bounds, including $\U^{(N)}$, $\V^{(N)}$, $\p^{(N)}$ and $\q^{(N)}$ in~(\ref{eq:linearize}), can be propagated along the computational graph.
This has been shown in detail in Appendix D of~\cite{liu2019certifying}.
Therefore, our method is also applicable to other network architectures, such as convolutional neural networks (CNN), residual networks (ResNet) and recurrent neural networks (RNN).
}

\section{Algorithms} \label{sec:certify}

\subsection{Robustness Guarantees by Polyhedral Envelope} \label{subsec:certify_bounds}

For an input point $\x$ with label $y \in [K]$, a sufficient condition to guarantee robustness is that the lower bounds of $\z'^{(N)}_y - \z'^{(N)}_i$ are positive for all $i \in [K]$.
\change{
Here, we use the \textit{elision of the last layer} introduced in~\cite{gowal2018effectiveness} to merge the subtraction of $\z'^{(N)}_y$ and $\z'^{(N)}_i$ with the last linear layer.
Therefore, we obtain the lower bound of $\z'^{(N)}_y - \z'^{(N)}_i$ as a whole: $\underline{\z'^{(N)}_y - \z'^{(N)}_i} \eqdef \U_i\x' + \p_i$.
}
Then, the sufficient condition to ensure robustness within a budget $\Set_{\epsilon}^{(p)}(\x)$ can be written as the following inequality:
\begin{equation}
\begin{aligned}
\underline{\z'^{(N)}_y - \z'^{(N)}_i} = \U_i\x' + \p_i \geq 0\,\; \ \forall i \in [K]\;.
\end{aligned} \label{eq:sufficient_condition}
\end{equation}

The constraint is trivial when $i = y$, so there are $(K - 1)$ such linear constraints, corresponding to $K - 1$ hyperplanes in the input space.
Within the adversarial budget, these hyperplanes provide a polyhedral envelope of the true decision boundary.
In the remainder of the paper, we use the term $d_{iy}$ to represent the distance between the input and the hyperplane defined in (\ref{eq:sufficient_condition}) and define $d_y = \min_{i \in [K], i \neq y} d_{iy}$ as the distance between the input and the polyhedral envelope's boundary.
The distance can be based on different $l_p$ norms, and $d_{iy} = 0$ when the input itself does not satisfy the inequality (\ref{eq:sufficient_condition}).
Since (\ref{eq:sufficient_condition}) is a sufficient condition for robustness given the adversarial budget $\Set^{(p)}_\epsilon(\x)$, it is guaranteed there is no adversarial example in the intersection of $\Set^{(p)}_\epsilon(\x)$ and the polytope defined in (\ref{eq:sufficient_condition}).

\change{The lemma below formalizes the vanilla case of our robustness certification, when there are no additional constraints on the input.}
We defer its proof to Appendix~\ref{subsec:proof_bound_thm} and call our method \textit{Polyhedral Envelope Certification} (PEC).

\change{
\begin{lemma}[PEC in Unconstrained Cases] \label{theory:bounds}
Given a model $f: \R^{n_1} \to [K]$ and an input point $\x$ with label $y$, let $\U$ and $\p$ in (\ref{eq:sufficient_condition}) be calculated using a predefined adversarial budget $\Set^{(p)}_\epsilon(\x)$. Then, there is no adversarial example inside an $l_p$ norm ball of radius $d$ centered around $\x$, with $d = \min\left\{\epsilon, d_y\right\}$, where $d_{iy} = \max\left\{0, \frac{\U_i\x + \p_i}{\|\U_i\|_q}\right\}$. $l_q$ is the dual norm of the $l_p$ norm, i.e., $\frac{1}{p} + \frac{1}{q} = 1$.
\end{lemma}
}

\begin{figure}
\centering
\includegraphics[scale = 0.4]{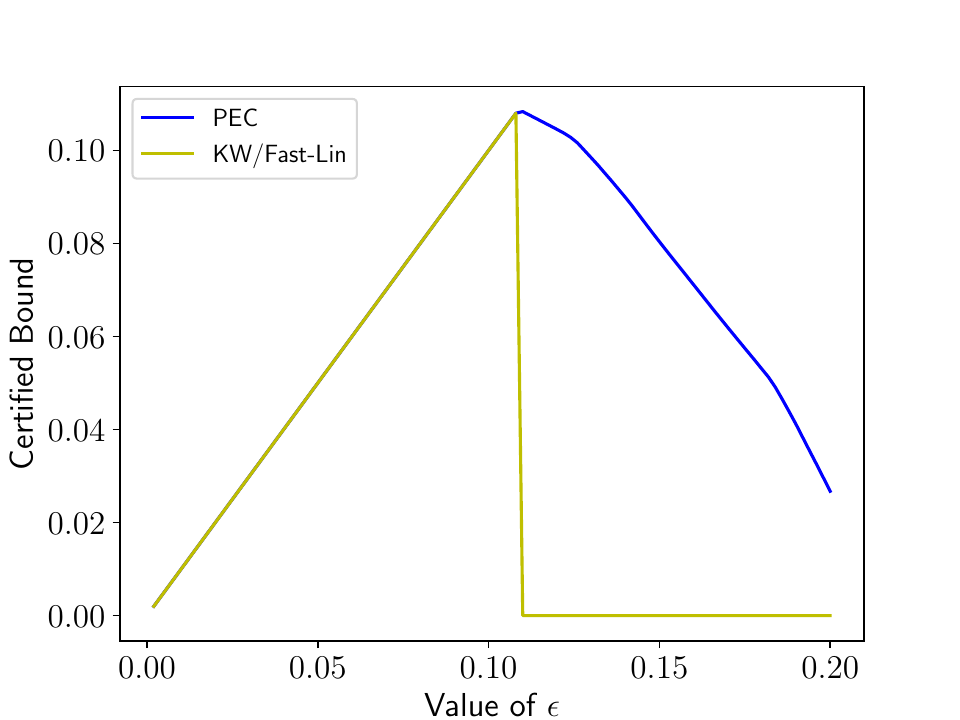}
\caption{Certified $l_\infty$-based bound of a randomly picked input by PEC and KW / Fast-Lin for different values of $\epsilon$. The model is the `FC1' model on MNIST trained by `MMR+at' in~\cite{croce2018provable}} \label{fig:two_phase_graph}
\vspace{-0.5cm}
\end{figure}

Based on Lemma \ref{theory:bounds}, when $\epsilon < d_y$, PEC has the same robustness guarantees as KW \cite{kolter2017provable}, Fast-Lin \cite{weng2018towards} and CROWN \cite{zhang2018efficient} using the same model linearization method.
When $0 < d_y < \epsilon$, KW / Fast-Lin / CROWN cannot certify the data point at all, while PEC still gives non-trivial robustness guarantees thanks to the geometric interpretability of the polyhedral envelope.
Figure~\ref{fig:two_phase_graph} compares the certified bounds of KW / Fast-Lin\footnote{In the case of ReLU networks, Fast-Lin and KW are algorithmically the same and yield the same robustness certification.} and PEC on a randomly picked input for different values of $\epsilon$ in the predefined adversarial budget.
\change{
We can clearly see the two-phase behavior of both methods.
In the second phase, unlike KW / Fast-Lin, PEC still provides a non-trivial certification bound.
}

\begin{figure}
\centering
\includegraphics[scale = 0.15]{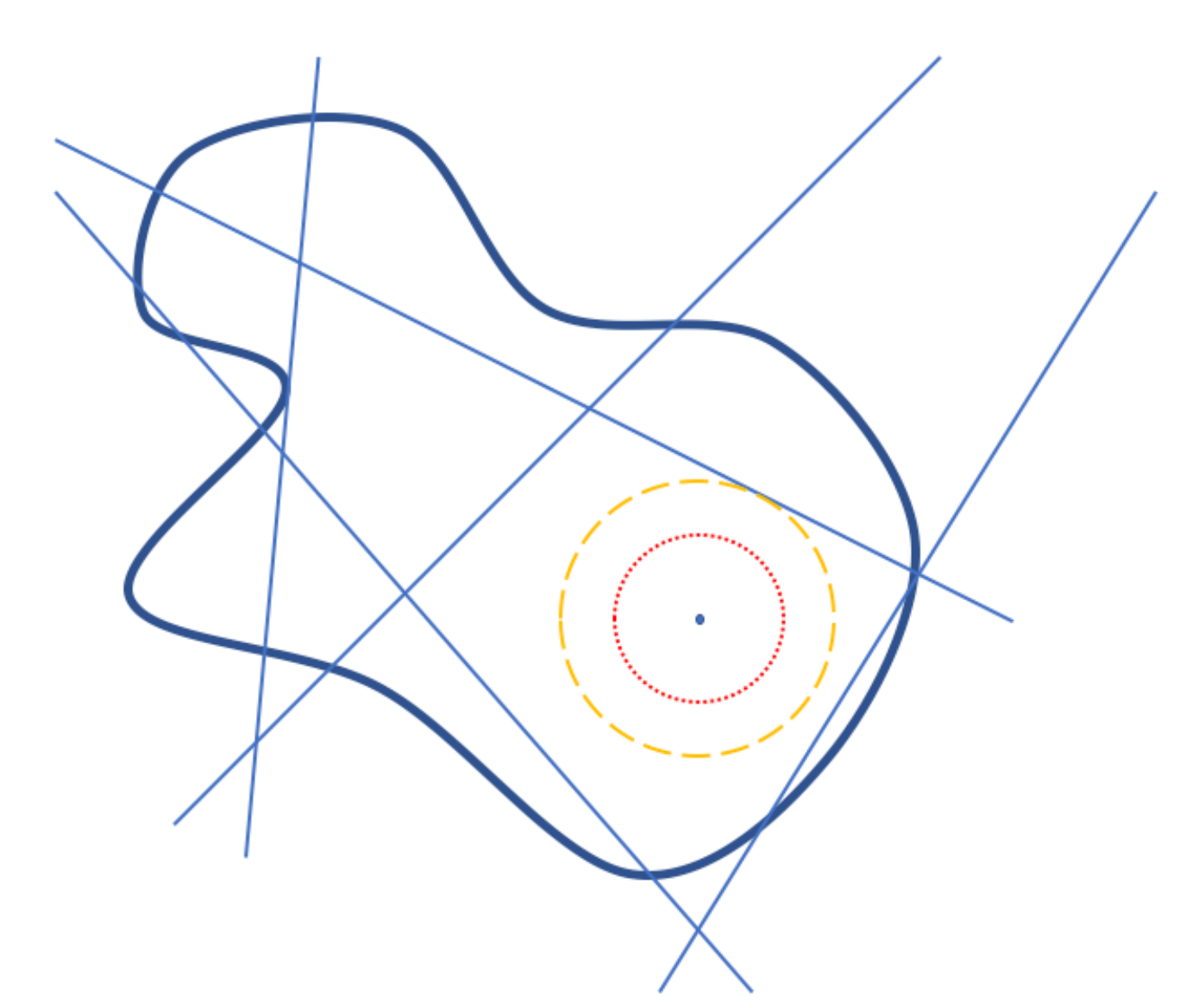} ~~~~~~
\includegraphics[scale = 0.15]{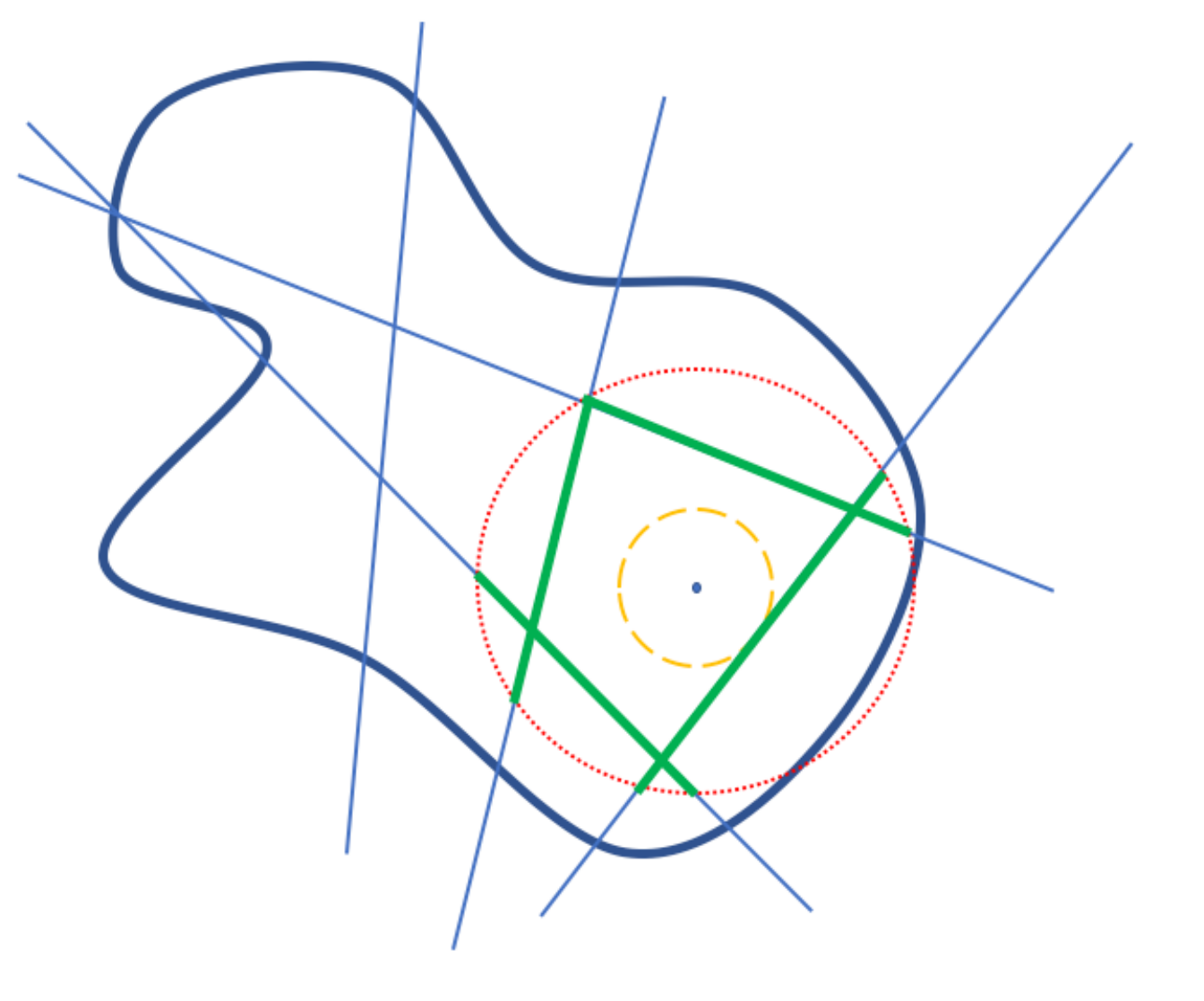}
\caption{2D sketch of decision boundary (dark blue bold lines), hyperplane defined by (\ref{eq:sufficient_condition}) (light blue lines), adversarial budget (red dotted circle), polyhedral envelope (green bold lines) in PEC. The distance between the input data and the hyperplanes is depicted by a yellow dashed circle. The left and right half correspond to the cases when $d_c$ is bigger and smaller than $\epsilon$, respectively.} \label{fig:two_phase_sketch}
\vspace{-0.2cm}
\end{figure}

Figure~\ref{fig:two_phase_sketch} shows a 2D sketch of the two phases mentioned above.
\change{
When $\epsilon$ is smaller than a threshold, as in the left half of the figure, the linear bounds in (\ref{eq:sufficient_condition}) are tight but only valid in a small region $\Set^{(p)}_\epsilon(\x)$.
Therefore, the certified robustness is $\epsilon$ at most.
When $\epsilon$ is bigger than this threshold, the linear bounds are valid in a larger region but becomes inevitably loose.
This is because the value of $d_{ic}$ monotonically decreases with the increase of $\epsilon$ for all model linearization methods.
This is depicted in the right half of the figure, where the distances between the input and the hyperplanes are smaller. The certified robustness is then $d_c$.
The hyperplane segments inside the adversarial budget (green bold lines) never exceed the decision boundary (dark blue bold lines), by definition of the polyhedral envelope.
The threshold here is the maximum certified bound, corresponding to the 'peak' of both curves in Figure~\ref{fig:two_phase_graph}.
We call this threshold \textit{the optimal value of $\epsilon$}.

To search for the optimal value of $\epsilon$,~\cite{kolter2017provable} uses Newton's method, which is an expensive second-order method.
\cite{weng2018towards, zhang2018efficient} use binary search to improve efficiency.
Thanks to the non-trivial certified bounds in the second phase, our proposed PEC can further accelerate their strategy.
During the search, when the value guess $\hat{\epsilon}$ is in the second phase, the vanilla binary search in Fast-Lin / CROWN can only conclude that the optimal value is smaller than $\hat{\epsilon}$.
In addition to this upper bound of the optimal value, PEC can output a non-trivial certified bound $d_c$, in which case we can also conclude that the optimal value is larger than $d_c$.
The tighter lower bound on the optimal value makes PEC need fewer steps to reach the required optimal value precision and thus accelerates the search.
We provide more detailed discussion and the pseudo code in Appendix~\ref{subsec:app_alg_search}.
}

In many applications, the input is constrained in a hypercube $[r^{(min)}, r^{(max)}]^{n_1}$.
For example, for images with normalized pixel intensities, an attacker will not perturb the image outside the hypercube $[0, 1]^{n_1}$.
Such constraint on the attacker allows us to ignore the regions outside the allowable input space, even if they are inside the adversarial budget $\Set^{(p)}_\epsilon(\x)$.

To obtain robustness guarantees in this scenario, we need to recalculate $d_{ic}$, which is now the distance between the input and the hyperplanes in (\ref{eq:sufficient_condition}) within the hypercube. 
The value of $d_{ic}$ is then the minimum of the following optimization problem 
\begin{equation}
\begin{aligned}
&~~~~~~~~~~~~~~~~~~~~~~ \min_\Delta \|\Delta\|_p & \\
& s.t.\ \ \ai\Delta + b \leq 0, ~~ \Delta^{(min)} \leq \Delta \leq \Delta^{(max)}
\end{aligned} \label{eq:bounded_constrained_problem}
\end{equation}
where, to simplify the notation, we define $\ai = \U_i$, $b = \U_i\x + \p$, $\Delta^{(min)} = r^{(min)} - \x$ and $\Delta^{(max)} = r^{(max)} - \x$.
When $b \leq 0$, the minimum is obviously $0$ as the optimal $\Delta$ is an all-zero vector.
In this case, either we cannot certify the input at all, or even the clean input is misclassified.
When $b > 0$, by H\"older's inequality, $\ai\Delta + b \geq -\|\Delta\|_p\|\ai\|_q + b$, with equality reached when $\Delta^p$ and $\ai^q$ are collinear.
Based on this, the optimal $\Delta$ of minimum $l_p$ norm to satisfy $\ai\Delta + b \leq 0$ is
\begin{equation}
\begin{aligned}
\widehat{\Delta}_i = -\frac{b}{\|\ai\|_q^q} \text{sign}(\ai_i) \left|\ai_i\right|^{\frac{q}{p}}\;,
\end{aligned} \label{eq:holder_optimal}
\end{equation}
where $\text{sign}(\cdot)$ returns $+1$ for positive numbers and $-1$ for negative numbers.

To satisfy the constraint $\Delta^{(min)} \leq \Delta \leq \Delta^{(max)}$, we use a greedy algorithm that approaches this goal progressively. That is, 
we first calculate the optimal $\widehat{\Delta}$ based on Equation~(\ref{eq:holder_optimal}) and check if the constraint $\Delta^{(min)} \leq \Delta \leq \Delta^{(max)}$ is satisfied.
For the elements where it is not, we clip their values within $[\Delta^{(min)}, \Delta^{(max)}]$ and keep them fixed.
We then optimize the remaining elements of $\Delta$ in the next iteration and repeat this process until the constraint is satisfied for all elements.
The pseudo-code is provided as Algorithm~\ref{alg:greedy} below and its optimality is guaranteed.

\begin{theory} \label{coro:optimal}
If the maximum number of iterations $I^{(max)}$ in Algorithm~\ref{alg:greedy} is large enough to satisfy $\Delta^{(min)} \leq \widehat{\Delta} \leq \Delta^{(max)}$ in Problem~\eqref{eq:bounded_constrained_problem}, then the output $\|\widehat{\Delta}\|_p$ is the optimum of Problem (\ref{eq:bounded_constrained_problem}), i.e., $d_{ic}$.
\end{theory}

We can use the primal-dual method to prove Theorem~\ref{coro:optimal}, which we defer to Appendix~\ref{subsec:proof_constrained_bounds}.
Once we have the value of $d_{ic}$ and thus $d_c$, the certified bound in this constrained case is then $\min\{\epsilon, d_c\}$, similar to Lemma~\ref{theory:bounds}.

\begin{algorithm}
\begin{algorithmic}[1]
\small
\STATE \textbf{Input:} $\x$, $\ai$, $b$, $\Delta^{(min)}$, $\Delta^{(max)}$ in~(\ref{eq:bounded_constrained_problem}) and maximum number of iterations allowed $I^{(max)}$
\STATE Set of fixed elements $\Set^{(f)} = \emptyset$
\STATE Iteration number $i = 0$
\STATE Calculate $\widehat{\Delta}$ according to (\ref{eq:holder_optimal})
\WHILE {$\Delta^{(min)} \leq \widehat{\Delta} \leq \Delta^{(max)}$ \text{not satisfied} and $i < I^{(max)}$}
    \STATE Violated entries $\Set^{(v)} = \{i | \widehat{\Delta}_i < \Delta^{(min)}_i\ \text{or}\ \widehat{\Delta}_i > \Delta^{(max)}_i\}$
    \STATE $\widehat{\Delta}_i = \text{clip}(\widehat{\Delta}_i, \text{min} = \Delta^{(min)}_i, \text{max} = \Delta^{(max)}_i), i \in \Set^{(v)}$
    \STATE $\Set^{(f)} = \Set^{(f)} \cup \Set^{(v)}$
    \STATE Update $\widehat{\Delta}$ according to (\ref{eq:holder_optimal}) with elements in $\Set^{(f)}$ fixed
    \STATE Update $i = i + 1$
\ENDWHILE
\STATE \textbf{Output:} $\|\widehat{\Delta}\|_p$
\end{algorithmic}
\caption{Greedy algorithm to solve Problem (\ref{eq:bounded_constrained_problem}).} \label{alg:greedy}
\end{algorithm}

\change{
If $I^{(max)}$ is set so small that the while-loop breaks with $\Delta^{(min)} \leq \widehat{\Delta} \leq \Delta^{(max)}$ unsatisfied, then the output of Algorithm~\ref{alg:greedy} is the upper bound of Problem~(\ref{eq:bounded_constrained_problem}), and thus we eventually get a suboptimal but still valid robustness guarantee.
\cite{croce2020minimally} solves the same problem when designing an attack and points out Algorithm~\ref{alg:greedy} will converge in $O(n_1 \log n_1)$ time.
We observed $I^{(max)} = 20$ to be sufficient to satisfy the condition in Theorem~\ref{coro:optimal}.
In practice, the while-loop breaks within 5 iterations in most cases, which means Algorithm~\ref{alg:greedy} introduces very little overhead.
}

\subsection{Geometry-Inspired Regularization} \label{subsec:regularization}

As in~\cite{croce2018provable}, we can incorporate our certified bounds in Theorem~\ref{coro:optimal} in the training process so as to obtain provably robust models.
To this end, we design a regularization term that encourages larger values of $d_{c}$.
We first introduce the \textit{signed distance} $\tilde{d}_{ic}$: when $d_{ic} > 0$, the clean input satisfies (\ref{eq:sufficient_condition}) and $\tilde{d}_{ic} = d_{ic}$; when $d_{ic} = 0$, the clean input does not satisfy (\ref{eq:sufficient_condition}) and there is no certified region; $\tilde{d}_{ic}$ in this case is a negative number whose absolute value is the distance between the input and the hyperplane defined in (\ref{eq:sufficient_condition}).
If the input is unconstrained, we have $\tilde{d}_{ic} = \frac{\U_i \x + \p_i}{\|\U_i\|_q}$.
Otherwise, following the notation of (\ref{eq:bounded_constrained_problem}), $\tilde{d}_{ic} = \text{sign}(b)\|\widehat{\Delta}\|_p$, where $\widehat{\Delta} = \argmin_\Delta \|\Delta\|_p,\; s.t.\; \ai\Delta + b = 0, \Delta^{(min)} \leq \Delta \leq \Delta^{(max)}$.
This problem can be solved by a greedy algorithm similar to the one in Section \ref{subsec:certify_bounds}.

Now, we sort $\{\tilde{d}_{ic}\}_{i = 0, i \neq c}^{K - 1}$ as $\tilde{d}_{j_0c} \leq \tilde{d}_{j_1c} \leq ... \leq \tilde{d}_{j_{K - 3}c} \leq \tilde{d}_{j_{K - 2}c}$ and then define the \textit{Polyhedral Envelope Regularization (PER)} term, based on the smallest $T$ distances, as
\begin{equation}
\begin{aligned}
\text{PER}(\x, \alpha, \gamma, T) = \gamma \sum_{i = 0}^{T - 1} \max \left(0, 1 - \frac{\tilde{d}_{j_ic}}{\alpha}\right)\;.
\end{aligned} \label{eq:per_def}
\end{equation}

Note that, following~\cite{croce2018provable}, to accelerate training, we take into account the smallest $T$ distances.
\change{
When $\tilde{d}_{j_ic} \geq \alpha$, the distance is considered large enough, so the corresponding term is zero and will not contribute to the gradient of the model parameters.
This avoids over-regularization and allows us to maintain accuracy on clean inputs.
}
In practice, we do not activate PER in the early training stages, when the model is not well trained and the corresponding polyhedral envelope is meaningless.
Such a `warm up' trick is commonly used in deep learning practice \cite{gotmare2018closer}.

We can further incorporate PER with adversarial training in a similar way to~\cite{croce2018provable}.
Here, the distance $\tilde{d}_{j_{ic}}$ in (\ref{eq:per_def}) is calculated between the polyhedral envelope and the adversarial example generated by PGD~\cite{madry2017towards} instead of the clean input.
Note that, the polyhedral envelope is the same in both cases because it only depends on the adversarial budget $\Set^{(p)}_\epsilon(\x)$.
We call this method \textit{PER+at}.

Calculating the polyhedral envelope is expensive in terms of both computation and memory because of the need to obtain linear bounds of the output logits.
\change{
We conduct a comprehensive complexity analysis in Section~\ref{sec:discussion}.
To prevent such a prohibitive computational and memory overhead,
we use the stochastic robust approximation in~\cite{wang2018mixtrain}.
For a mini-batch of size $B$, we only calculate the PER or PER+at regularization term for $B' < B$ instances randomly sub-sampled from this mini-batch.
Each instances in the mini-batch has the same probability to be sampled.
}
\cite{moosavi2017universal} empirically observed the geometric correlation of high-dimensional decision boundaries near the data manifold.
Although this finding is based on regularly trained models, we find it also holds for models trained by PER / PER+at: in practice, a $B'$ much smaller than $B$ provides a good approximation of the full-batch regularization.

The full pipeline of PER+at method is demonstrated as Algorithm~\ref{alg:perat}.
$\mathcal{D}$ and $\loss$ represent the dataset and the loss function, respectively.

\begin{algorithm}
\begin{algorithmic}[1]
\small
\STATE \textbf{Input:} $\mathcal{D}$, $\gamma$, $\alpha$, $T$, $B$, $B'$
\STATE Sample $(\mathbf{X}, \mathbf{y}) \in (\R^{B \times m}, [K]^{B})$ from the dataset $\mathcal{D}$.
\STATE Subsample $(\mathbf{X}_{s}, \mathbf{y}_{s}) \in (\R^{B' \times m}, [K]^{B'})$ from the minibatch.
\STATE Using model linearization to calculate $\U$ and $\p$ in Equation~(\ref{eq:linearize}) for each instance in $(\mathbf{X}_s, \mathbf{y}_s)$.
\STATE Using PGD attack to generate adversarial examples $(\mathbf{X}', \mathbf{y}')$ of the whole mini-batch, including the subsamples.
\STATE Calculate PER regularization term based on linearization $\U$, $\p$ and input $(\mathbf{X}'_s, \mathbf{y}'_s)$ using Algorithm~\ref{alg:greedy}.
\STATE The final loss is $\frac{1}{2} (\loss(\mathbf{X}, \mathbf{y}) + \loss(\mathbf{X}', \mathbf{y}')) + \mathrm{PER}(\mathbf{X}'_s, \alpha, \gamma, T)$.
\STATE Back-propagation and update model parameters.
\end{algorithmic}
\caption{Full pipeline of PER+at method} \label{alg:perat}
\end{algorithm}


\section{Experiments} \label{sec:experiments}

\begin{table*}[!ht]
\centering
\small
\begin{tabular}{p{2.2cm}p{1.3cm}<{\raggedleft\arraybackslash}p{1.3cm}<{\raggedleft\arraybackslash}p{1.3cm}<{\raggedleft\arraybackslash}p{1.3cm}<{\raggedleft\arraybackslash}p{1.3cm}<{\raggedleft\arraybackslash}p{1.3cm}<{\raggedleft\arraybackslash}p{1.3cm}<{\raggedleft\arraybackslash}p{1.3cm}<{\raggedleft\arraybackslash}}
\Xhline{4\arrayrulewidth}
Methods & CTE ~~~ (\%) & PGD ~~~ (\%) & CRE Lin (\%) & CRE IBP (\%) & CRE MIP (\%) & ACB Lin & ACB IBP & ACB PEC \\
\hline
\\[-0.2cm]
& \multicolumn{8}{c}{\textbf{MNIST - FC1, ReLU, $l_\infty$, $\epsilon = 0.1$}} \\[0.1cm]
plain               & 1.99     & 98.37    & 100.00    & 100.00    & 100.00     & 0.0000     & 0.0000     & 0.0000     \\
at                  & 1.42     & 9.00     & 97.94     & 100.00    & 100.00     & 0.0021     & 0.0000     & 0.0099     \\
KW                  & 2.26     & 8.59     & 12.91     & 69.20     & 10.90      & 0.0871     & 0.0308     & 0.0928     \\
IBP                 & 1.65     & 9.67     & 87.27     & \two{15.20} & 12.36      & 0.0127     & \two{0.0848} & 0.0705     \\
C-IBP               & 1.98     & 9.50     & 67.39     & \one{14.45} & 11.39      & 0.0326     & \one{0.0855} & 0.0800     \\
MMR                 & 2.11     & 17.82    & 33.75     & 99.88     & 24.90      & 0.0663     & 0.0001     & 0.0832     \\
MMR+at              & 2.04     & 10.39    & 17.64     & 95.09     & 14.10      & 0.0824     & 0.0049     & 0.0905     \\
\textbf{C-PER}      & \two{1.60}& \two{7.45} & \one{11.71} & 92.89     & \one{7.69}   & \one{0.0883} & 0.0071     & \one{0.0935} \\
\textbf{C-PER+at}   & 1.81     & 7.73     & 12.90     & 99.90     & 8.22       & 0.0871     & 0.0001     & 0.0925     \\
\textbf{I-PER}      & \two{1.60} & \one{6.28} & \two{11.96} & 93.33     & \two{8.10}   & \two{0.0880} & 0.0067     & \two{0.0934} \\
\textbf{I-PER+at}   & \one{1.54} & 7.15     & 13.96     & 98.55     & 8.48       & 0.0868     & 0.0014     & 0.0927     \\
\hline
\\[-0.2cm]
& \multicolumn{8}{c}{\textbf{MNIST - CNN, ReLU, $l_\infty$, $\epsilon = 0.1$}} \\[0.1cm]
plain               & 1.28     & 85.75    & 100.00   & 100.00   & 100.00     & 0.0000     & 0.0000     & 0.0000     \\
at                  & 1.02     & 4.75     & 91.91    & 100.00   & 100.00     & 0.0081     & 0.0000     & 0.0189     \\
KW                  & 1.21     & 3.03     & \one{4.44} & 100.00   & 4.40       & \one{0.0956} & 0.0000     & \one{0.0971} \\
IBP                 & 1.51     & 4.43     & 23.89    & \two{8.13} & 5.23       & 0.0761     & \two{0.0919} & 0.0872     \\
C-IBP               & 1.85     & 4.28     & 10.72    & \one{6.91} & 4.83       & 0.0893     & \one{0.0931} & 0.0928     \\
MMR                 & 1.65     & 6.07     & 11.56    & 100.00   & 6.10       & 0.0884     & 0.0000     & 0.0928     \\
MMR+at              & 1.19     & 3.35     & 9.49     & 100.00   & 3.60       & 0.0905     & 0.0000     & 0.0939     \\
\textbf{C-PER}      & 1.44     & 3.44     & 5.13     & 100.00   & 3.62       & 0.0949     & 0.0000     & 0.0965     \\
\textbf{C-PER+at}   & \one{0.50} & \two{2.02} & 4.85     & 100.00   & \two{2.21}   & 0.0952     & 0.0000     & 0.0969     \\
\textbf{I-PER}      & 1.03     & 2.40     & 4.64     & 99.55    & 2.52       & \two{0.0954} & 0.0004     & 0.0967     \\
\textbf{I-PER+at}   & \two{0.48} & \one{1.29} & \two{4.61} & 99.94    & \one{1.47}   & \two{0.0954} & 0.0001     & \one{0.0971} \\
\hline
\\[-0.2cm]
& \multicolumn{8}{c}{\textbf{CIFAR10 - CNN, ReLU, $l_\infty$, $\epsilon = 2 / 255$}} \\[0.1cm]
plain               & 24.62     & 86.29     & 100.00    & 100.00    & 100.00     & 0.0000     & 0.0000     & 0.0000     \\
at                  & 27.04     & 48.53     & 85.36     & 100.00    & 88.50      & 0.0011     & 0.0000     & 0.0015     \\
KW                  & 39.27     & 46.60     & \one{53.81} & 99.98     & \two{48.00}  & \one{0.0036} & 0.0000     & \one{0.0040} \\
IBP                 & 46.74     & 56.38     & 61.81     & \one{67.58} & 58.80      & 0.0030     & \one{0.0025} & 0.0034     \\
C-IBP               & 58.32     & 63.56     & 66.28     & \two{69.10} & 65.44      & 0.0026     & \two{0.0024} & 0.0029     \\
MMR                 & 34.59     & 57.17     & 69.28     & 100.00    & 61.00      & 0.0024     & 0.0000     & 0.0032     \\
MMR+at              & 35.36     & 49.27     & 59.91     & 100.00    & 54.20      & 0.0031     & 0.0000     & 0.0037     \\
\textbf{C-PER}      & 39.21     & 50.98     & 57.45     & 99.98     & 52.70      & 0.0033     & 0.0000     & 0.0038     \\
\textbf{C-PER+at}   & \two{28.87} & \one{43.55} & \two{56.59} & 100.00    & 48.43      & \two{0.0034} & 0.0000     & \one{0.0040} \\
\textbf{I-PER}      & 29.34     & 51.54     & 64.34     & 99.98     & 54.87      & 0.0028     & 0.0000     & 0.0036     \\
\textbf{I-PER+at}   & \one{26.66} & \two{43.35} & 57.72     & 100.00    & \one{47.87}  & 0.0033     & 0.0000     & \one{0.0040} \\
\Xhline{4\arrayrulewidth}
\end{tabular}
\caption{\change{Full results of 11 training schemes and 8 evaluation schemes for ReLU networks under $l_\infty$ attacks. The best and the second best results among provably robust training methods (plain and at excluded) are bold. In addition, the best results are underlined.}} \label{tbl:main}
\vspace{-0.4cm}
\end{table*}

To validate the theorem and algorithms above, we conducted several experiments on two popular image classification benchmarks: MNIST and CIFAR10.
\change{
Each of these experiments can be completed on a single NVIDIA TITAN XP GPU machine of 12GB memory within several hours.
Our code and checkpoints are publicly available at \href{https://github.com/liuchen11/PolyEnvelope}{https://github.com/liuchen11/PolyEnvelope}.
}

\subsection{Training and Certifying ReLU Networks} \label{subsec:exp_relu}

We first demonstrate the benefits of our approach over existing training and certification methods under the same computational complexity.
To this end, we use the same model architectures as in~\cite{croce2018provable, kolter2017provable}:
\textbf{FC1}, which is a fully-connected network with one hidden layer of 1024 neurons; and
\textbf{CNN}, which has two convolutional layers followed by two fully-connected layers. For this set of experiments, all activation functions are ReLU.

When it comes to training, we consider 7 baselines, including plain training (plain), adversarial training (at) \cite{madry2017towards}, KW \cite{kolter2017provable}, IBP \cite{gowal2018effectiveness},
CROWN-IBP \cite{zhang2019towards}, MMR and MMR plus adversarial training (MMR + at) \cite{croce2018provable}.
We denote our method as C-PER, C-PER+at when we use CROWN-style model linearization for PER and PER+at, respectively, and as I-PER and I-PER+at when using IBP-inspired model linearization.
We do not compare randomized smoothing \cite{cohen2019certified, salman2019provably} or layerwise training \cite{balunovic2020adversarial}.
This is because the certified bounds of randomized smoothing are not exact but probabilistic, and layerwise training has significant computational overhead.\footnote{\label{note2}
For CNN models,~\cite{balunovic2020adversarial} trains 200 epochs for each layer and 800 epochs in total, while the other baselines use only 100 epochs. If we reduce the training epochs of each layer to 25 epochs, the model does not converge well. For FC1 models,~\cite{balunovic2020adversarial} is the same as KW, because there is only one hidden layer.}
For fair comparison, we use the same adversarial budget in both the training and the test phases.

To evaluate the models' performance on the test set, we first report the clean test error (CTE) and the empirical robust error against PGD (PGD).
Based on the discussions in Section~\ref{subsec:certify_bounds}, KW, Fast-Lin and PEC have the same certified robust error, which is the proportion of the input data whose certified regions are smaller than the adversarial budget.
Therefore, for these three methods, we report the certified robust error as CRE Lin.
We also report the certified robust error by IBP \cite{gowal2018effectiveness}.
For $l_\infty$ robustness, we use a complete certifier called MIPVerify \cite{tjeng2017evaluating} to calculate the exact robust error, denoted by CRE MIP. \footnote{MIPVerify is available on \href{https://github.com/vtjeng/MIPVerify.jl}{https://github.com/vtjeng/MIPVerify.jl}}
In addition, we calculate the average certified bound obtained by Fast-Lin / KW (ACB Lin)\footnote{Fast-Lin and KW is algorithmically the same in ReLU networks}, IBP (ACB IBP) and PEC (ACB PEC).
Note that the average certified bound here is from the \textit{one-shot} certifier, i.e., without searching for the optimal adversarial budget.
We do not report the certified bound obtained by MMR \cite{croce2018provable}, because, in practice, it only gives trivial results.
As a matter of fact,~\cite{croce2018provable} emphasize their training method and report certification results using only KW and MIP.

\change{
We use the same adversarial budgets and model architectures as~\cite{croce2018provable} and thus directly download the KW, MMR and MMR+at models from the checkpoints provided online.\footnote{\href{https://github.com/max-andr/provable-robustness-max-linear-regions}{https://github.com/max-andr/provable-robustness-max-linear-regions}.}
For IBP and CROWN-IBP (C-IBP), we use the same hyper-parameter settings as~\cite{zhang2019towards} except that we align the training duration to other methods and the use stochastic robustness approximation of Section~\ref{subsec:regularization} to reduce the computational and memory consumption.
For CNN models, we use the \textit{warm up} trick consisting of performing adversarial training before adding our PER or PER+at regularization term.
The running time overhead of pre-training is negligible compared with computing the regularization term.
We train all models for 100 epochs and provide the detailed hyper-parameter settings in Appendix~\ref{subsec:app_experiment_details}.
}

\begin{figure}
\centering
\includegraphics[scale = 0.5]{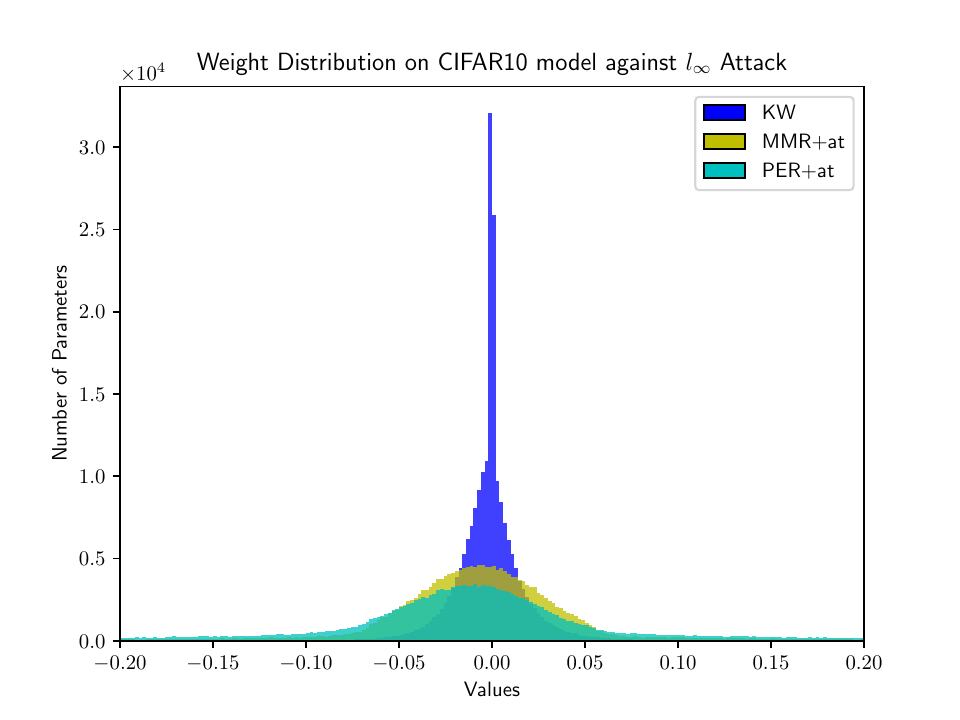}
\caption{Parameter value distributions of CIFAR10 models trained against $l_\infty$ attacks. The Euclidean norms of KW, MMR+at, PER+at models against $l_\infty$ attack are 18.08, 38.36 and 94.63 respectively, which evidences that the KW model is over-regularized while our PER model best preserves the model capacity.} \label{fig:param_dist}
\vspace{-0.4cm}
\end{figure}

We constrain the attacker to perturb the images within $[0, 1]^{n_1}$, and the full results for $l_\infty$ attacks are summarized in Table~\ref{tbl:main}.
The results of $l_2$ attacks are demonstrated in Table~\ref{tbl:main_l2} of Appendix~\ref{subsubsec:app_relu_l2}.
\change{
For $l_\infty$ attacks, our (C/I)-PER or (C/I)-PER+at achieve the best certified accuracy, calculated by the complete certifier (CRE MIP), in all cases.
For $l_2$ attacks, they also achieve the best estimated certified accuracy, calculated by the Fast-Lin / KW / PEC certifier (CRE Lin), in all cases.
}
In addition, the performance of I-PER and I-PER+at is on par with that of C-PER and C-PER+at, which illustrates that our framework is not sensitive to the tightness of the underlying model linearization method and thus generally applicable.

As observed in previous work \cite{raghunathan2018certified}, different incomplete certifiers are complementary; IBP is only able to certify IBP-trained models and has worse certification results on other models.
\minorchange{
For the training methods other than IBP and C-IBP, we notice big gaps between the true robustness (CRE MIP) and the IBP certified robustness (CRE IBP).
This is because IBP and C-IBP solve a different optimization problem from the other methods.
Specifically, IBP and C-IBP do not make any approximation of the activation function, they only utilize the monotonicity of the activation function to propagate the bounds.
However, all the other methods use linear approximations to bound the outputs of the activation functions.
}
\change{
We also note that the stochastic robustness approximation greatly hurts the performance of IBP and C-IBP on CIFAR10.
However, the result reported in~\cite{zhang2018efficient} without stochastic robustness approximation on the same architecture is still worse than our method.\footnote{The DM-small model in~\cite{zhang2018efficient} yields a certified robust error of $52.57\%$ on CIFAR10 when $\epsilon = 2/255$.}
}
Consistently with Section~\ref{subsec:certify_bounds}, our geometry-inspired PEC has better average certified bounds than Fast-Lin / KW given the same adversarial budget.
For example, on the CIFAR10 model against $l_\infty$ attack, $10\% - 20\%$ of the test points are not certified by Fast-Lin / KW but have non-trivial bounds with PEC.

When compared with KW, our methods, especially PER+at, have much better clean test accuracy.
In other words, a model trained by (C/I)-PER+at is not as over-regularized as other training methods for provable robustness.
Figure~\ref{fig:param_dist} shows the distribution of parameter values of KW, MMR+at, C-PER+at models on CIFAR10 against $l_\infty$ attacks.
The results of CIFAR10 models against the $l_2$ attack are shown in Figure~\ref{fig:param_dist_app} of Appendix~\ref{subsec:app_param_value}.
As we can see, the parameters of C-PER+at models have much larger norms than KW and MMR+at, whose parameters are more sparse.
The norms of the model parameters indicate the model capacity~\cite{neyshabur2014search,neyshabur2017exploring}, so C-PER+at models better preserve the model capacity.

\begin{figure}[h]
\centering
\includegraphics[scale = 0.45]{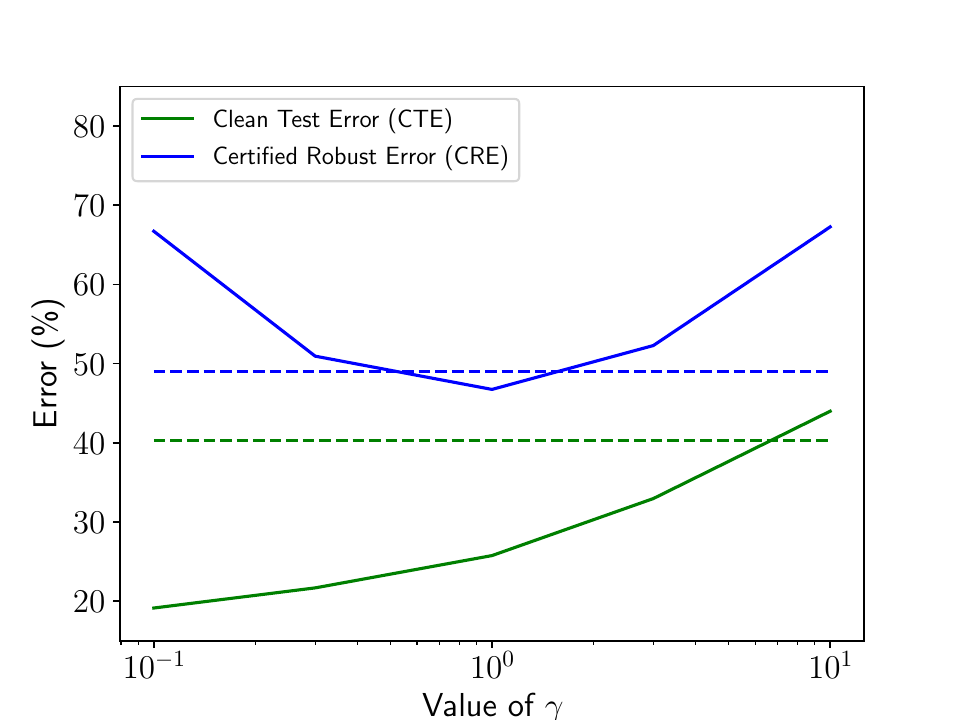}
\caption{CTE and CRE for different values of $\gamma$ in C-PER+at to show their trade-off. The results of KW, for reference, are the horizontal dashed lines. The optimal value of $\gamma$ for C-PER+at is $1.0$, with both CTE and CRE better than KW.} \label{fig:tradeoff}
\vspace{-0.4cm}
\end{figure}

The better performance of (C/I)-PER+at over (C/I)-PER, and of MMR+at over MMR, evidences the benefits of augmenting the training data with the adversarial examples.
\change{
However, this strategy is only compatible with methods that rely on estimating the distance between the data point and the decision boundary, and thus cannot be combined with methods such as KW.
Adding a loss term on the adversarial examples to the loss objective of KW yields a performance between adversarial training and KW.
For example, if we optimize the sum of loss objectives of KW and PGD in MNIST - CNN $l_\infty$ cases, the robust error against PGD of the resulting model is $3.64\%$, the provably robust error by Fast-Lin (CRE Lin) is $8.12\%$.
In other words, such combinations only lead to mixed performance and are weaker than KW in terms of provable robustness.
}

\begin{table*}[h]
\centering
\small
\begin{tabular}{p{2.2cm}p{1.4cm}<{\raggedleft\arraybackslash}p{1.4cm}<{\raggedleft\arraybackslash}p{1.4cm}<{\raggedleft\arraybackslash}p{1.4cm}<{\raggedleft\arraybackslash}p{1.4cm}<{\raggedleft\arraybackslash}p{1.4cm}<{\raggedleft\arraybackslash}p{1.4cm}<{\raggedleft\arraybackslash}}
\Xhline{4\arrayrulewidth}
Methods & CTE ~~~ (\%) & PGD ~~~ (\%) & CRE CRO (\%) & CRE IBP (\%) & ACB CRO & ACB IBP & ACB PEC \\
\hline
\\[-0.2cm]
& \multicolumn{7}{c}{\textbf{MNIST - FC1, Sigmoid, $l_\infty$, $\epsilon = 0.1$}} \\[0.1cm]
plain              & 2.04     & 97.80     & 100.00     & 100.00     & 0.0000     & 0.0000     & 0.0000 \\
at                 & 1.78     & 10.05     & 98.52      & 100.00     & 0.0015     & 0.0000     & 0.0055 \\
IBP                & 2.06     & 10.58     & 44.14      & 13.65      & 0.0559     & 0.0863     & 0.0846 \\
C-IBP              & 2.88     &  9.83     & 26.04      & \one{12.51}  & 0.0740     & \one{0.0875} & 0.0886 \\
\textbf{C-PER}     & \one{1.97} &  7.55     & 12.15      & 84.76      & 0.0879     & 0.0152     & \one{0.0930} \\
\textbf{C-PER+at}  & 2.16     &  \one{7.12} & \one{11.87}  & 88.06      & \one{0.0881} & 0.0119     & 0.0927 \\
\textbf{I-PER}     & 2.15     &  8.35     & 12.79      & 86.99      & 0.0872     & 0.0130     & 0.0926 \\
\textbf{I-PER+at}  & 2.45     &  8.05     & 12.36      & 88.94      & 0.0876     & 0.0111     & 0.0923 \\
\hline
\\[-0.2cm]
& \multicolumn{7}{c}{\textbf{MNIST - FC1, Tanh, $l_\infty$, $\epsilon = 0.1$}} \\[0.1cm]
plain              & 2.00     & 97.80     & 100.00     & 100.00     & 0.0000     & 0.0000     & 0.0000 \\
at                 & 1.28     &  8.89     &  99.98     & 100.00     & 0.0000     & 0.0000     & 0.0001 \\
IBP                & \one{2.04} &  9.84     &  31.81     &  13.02     & 0.0682     & 0.0870     & 0.0864 \\
C-IBP              & 2.75     &  9.57     &  20.10     &  \one{11.80} & 0.0799     & \one{0.0882} & 0.0894 \\
\textbf{C-PER}     & 2.19     &  7.71     &  11.55     &  57.81     & 0.0885     & 0.0422     & \one{0.0934}\\
\textbf{C-PER+at}  & 2.30     &  \one{7.45} &  \one{11.39} &  56.74     & \one{0.0886} & 0.0433     & 0.0930 \\
\textbf{I-PER}     & 2.21     &  8.51     &  12.23     &  55.53     & 0.0878     & 0.0445     & 0.0929 \\
\textbf{I-PER+at}  & 2.46     &  7.87     &  12.04     &  66.04     & 0.0880     & 0.0340     & 0.0929 \\
\Xhline{4\arrayrulewidth}
\end{tabular}
\caption{\change{Full results of 8 training schemes and 7 evaluation schemes for sigmoid and tanh networks under $l_\infty$ attacks. The best results among provably robust training methods (plain and at excluded) are bold and underlined.}} \label{tbl:nonrelu}
\vspace{-0.2cm}
\end{table*}

To demonstrate the trade-off between clean test error and certified robust error, we evaluate our approach with different regularizer strength $\gamma$ in Equation (\ref{eq:per_def}).
Figure~\ref{fig:tradeoff} shows the example of C-PER+at in the $l_2$ case for CIFAR10.
When $\gamma$ is small, the PER term has little influence on training, and C-PER+at becomes similar to adversarial training (at).
It has low clean test error but very high certified robust error.
As $\gamma$ grows, the model is increasingly regularized towards large polyhedral envelopes, which inevitably hurts the performance on the clean input.
By contrast, the certified robust error first decreases and then increases.
This is because training is numerically more difficult when $\gamma$ is too large and the model is over-regularized.
The results of KW are shown as horizontal dashed lines for comparison.
We can see that C-PER+at is in general less over-regularized than KW, with much lower clean test error for the same certified robust error.

\minorchange{
To more comprehensively study the performance of our proposed methods, we conduct experiments on larger adversarial budgets.
We compare our proposed C-PER+at and I-PER+at with C-IBP~\cite{zhang2019towards}.
}
\change{
Here, we use the \minorchange{\textit{model architecture E}} from~\cite{zhang2019towards}, consisting of 3 convolutional layers and 2 fully connected layers.\footnote{Models available on \href{https://github.com/huanzhang12/CROWN-IBP}{https://github.com/huanzhang12/CROWN-IBP}}
We focus on MNIST, where the model can achieve a decent certified robust accuracy under large adversarial budgets, and set $\epsilon$ to be $0.1$, $0.2$, $0.3$ and $0.4$.
\minorchange{
For C-IBP, we directly use the publicly available checkpoints from~\cite{zhang2019towards}.
For (C/I)-PER+at, we use the same settings as the ones in Table~\ref{tbl:main} except for the change of adversarial budget.
Table~\ref{tab:big_budget} compares the exact certified robust error by MIP (CRE MIP) among the different training methods.
}
The results show that our proposed (C/I)-PER+at yields better results than C-IBP when $\epsilon$ is $0.1$ and $0.2$ but underperforms it when $\epsilon$ is $0.3$ and $0.4$.
This phenomenon indicates that our method is more suitable when the adversarial budget is relatively small.
This arises from the trade-off between model linearization and interval bound propagation.
\minorchange{
When the adversarial budget is small, the lower and upper bounds of most ReLU neurons have smaller gaps and are either both negative or both positive.
Such neurons can be considered linear and make the model linearization methods, which (C/I)-PER+at is based on, more accurate.
With a more accurate bound in the loss function, (C/I)-PER+at outperforms C-IBP, which accumulates the estimation error faster layerwisely~\cite{zhang2018efficient}.
By contrast, a recent study~\cite{lee2021loss} shows that the loss landscape of training methods based on model linearization is less smooth than the ones of IBP and C-IBP.
In addition,~\cite{liu2020loss} demonstrates that with the increase of the adversarial budget the loss landscape becomes even more challenging.
As a result, when using (C/I)-PER+at with a large adversarial budget, the optimizers cannot find a good minimum.
This makes (C/I)-PER+at underperform C-IBP.
}
}

\begin{table}[h]
\centering
\small
\begin{tabular}{p{1.8cm}p{0.90cm}<{\raggedleft\arraybackslash}p{0.90cm}<{\raggedleft\arraybackslash}p{0.90cm}<{\raggedleft\arraybackslash}p{0.90cm}<{\raggedleft\arraybackslash}}
\Xhline{4\arrayrulewidth}
\\[-0.2cm]
Value of $\epsilon$ & 0.1 & 0.2 & 0.3 & 0.4 \\[0.1cm]
\hline
\\[-0.2cm]
\textbf{C-IBP}    & 3.90 & 7.25 & \one{11.28} & \one{18.58} \\
\textbf{C-PER+at} & \one{3.52} & 7.09 & 11.34 & 20.12 \\
\textbf{I-PER+at} & 3.58 & \one{7.05} & 11.42 & 21.02 \\
\Xhline{4\arrayrulewidth}
\end{tabular}
\caption{\change{Exact certified robust error by MIP (CRE MIP) of different methods under different sizes of the $l_\infty$ adversarial budget on MNIST. The best results are bold and underlined.}}
\label{tab:big_budget}
\vspace{-0.2cm}
\end{table}

\subsection{Training and Certifying Non-ReLU Networks}

To validate our method's applicability to non-ReLU networks, we replace the ReLU function in FC1 models with either sigmoid or tanh functions.
MMR and MMR+at are no longer applicable here, because they only support piece-wise linear activation functions.
MIPVerify does not support sigmoid or tanh functions neither, since it works only on ReLU networks.
While~\cite{wong2018scaling} claims that their methods apply to non-ReLU networks, their main contribution is rather the extension of KW to a broader set of network architectures, and their public code\footnote{Repository: \href{https://github.com/locuslab/convex_adversarial}{https://github.com/locuslab/convex\_adversarial}} does not support non-ReLU activations.
For evaluation, we replace Fast-Lin and KW with CROWN~\cite{zhang2018efficient} and thus report its certified robust error (CRE CRO) and average certified bound (ACB CRO).
We use the model linearization method in Appendix~\ref{subsec:app_linearize_activation} for (C/I)-PER and (C/I)-PER+at during training, which is slightly different from CROWN.
This is because we need an analytical form of the linearization in order to calculate the model parameters' gradients.
When we certify models using CROWN, the model linearization method in~\cite{zhang2018efficient} is used because it is tighter.

The results on $l_\infty$ cases are shown in Table~\ref{tbl:nonrelu} and the ones on $l_2$ cases are demonstrated in Table~\ref{tbl:nonrelu_l2} of Appendix~\ref{subsubsec:app_nonrelu_l2}.
Similar to the ReLU networks in Section~\ref{subsec:exp_relu}, our (C/I)-PER and (C/I)-PER+at methods have the best performance in all cases, in terms of both certified robust error and average certified bound.
IBP can only certify IBP-trained models well and has significantly worse results on other models.

\subsection{Optimal Adversarial Budget} \label{subsec:optimal_budget}

To obtain the biggest certified bound based on the current model linearization method, we need to search for the optimal value of $\epsilon$, i.e., the peak in Figure~\ref{fig:two_phase_graph}.
KW \cite{kolter2017provable} uses Newton's method to solve a constrained optimization problem, which is expensive.
Fast-Lin and CROWN \cite{weng2018towards, zhang2018efficient} apply a binary search strategy to find the optimal $\epsilon$.
Based on the discussion in Section~\ref{subsec:certify_bounds}, the optimal adversarial budget for a data point is also its optimal certified bound.

To validate the claim in Section~\ref{subsec:certify_bounds} that PEC can find the optimal adversarial budget faster than Fast-Lin / CROWN, we compare the average number of iterations needed to find the optimal value given a predefined precision requirement $\epsilon_\Delta$.
Using $\underline{\epsilon}$ and $\bar{\epsilon}$ to define the initial lower and upper estimates of the optimal value, we then need $\ceil{\log_2 \frac{\bar{\epsilon} - \underline{\epsilon}}{\epsilon_\Delta}}$ steps of bound calculations to obtain the optimal value by binary search in Fast-Lin / CROWN.
By contrast, the number of bound calculations needed by PEC is smaller and depends on the model to certify, because the partial certified bounds obtained by PEC indicate tighter lower bounds of the optimal adversarial budget.
Our experimental results are based on Algorithm~\ref{alg:app_search} presented in Appendix~\ref{subsec:app_alg_search}.

\begin{table*}[h]
\centering
\small
\begin{tabular}{p{1.4cm}:p{1cm}<{\centering}p{1.15cm}<{\raggedleft\arraybackslash}p{1.15cm}<{\centering}:p{1cm}<{\centering}p{1.15cm}<{\raggedleft\arraybackslash}p{1.15cm}<{\centering}:p{1cm}<{\centering}p{1.15cm}<{\raggedleft\arraybackslash}p{1.15cm}<{\centering}}
\Xhline{4\arrayrulewidth}
\multirow{3}{*}{Methods} & \multicolumn{3}{c}{} & \multicolumn{3}{:c}{} & \multicolumn{3}{:c}{} \\
& \multicolumn{3}{c}{\textbf{MNIST-FC1, ReLU, $l_\infty$}} & \multicolumn{3}{:c}{\textbf{MNIST-CNN, ReLU, $l_\infty$}} & \multicolumn{3}{:c}{\textbf{CIFAR10-CNN, ReLU, $l_\infty$}} \\[0.1cm]
& {\normalsize T\textsubscript{Lin}} & {\normalsize T\textsubscript{PEC}} & {\normalsize $\frac{\text{T}_{\text{PEC}}}{\text{T}_{\text{Lin}}}$} & {\normalsize T\textsubscript{Lin}} & {\normalsize T\textsubscript{PEC}} & {\normalsize $\frac{\text{T}_{\text{PEC}}}{\text{T}_{\text{Lin}}}$} & {\normalsize T\textsubscript{Lin}} & {\normalsize T\textsubscript{PEC}} & {\normalsize $\frac{\text{T}_{\text{PEC}}}{\text{T}_{\text{Lin}}}$} \\[0.2cm]
\hline
plain    & \multirow{7}{*}{12} & 9.85  & 0.8207 & \multirow{7}{*}{12} & 10.56 & 0.8804 & \multirow{7}{*}{10} & 9.33  & 0.9331 \\
at       &                     & 10.77 & 0.8972 &                     & 11.39 & 0.9489 &                     & 9.12  & 0.9128 \\
KW       &                     & 8.48  & 0.7066 &                     & 11.61 & 0.9674 &                     & 8.43  & 0.8432 \\
MMR      &                     & 8.04  & 0.6703 &                     & 10.68 & 0.8897 &                     & 8.05  & 0.8053 \\
MMR+at   &                     & 7.68  & 0.6402 &                     & 11.22 & 0.9351 &                     & 8.45  & 0.8450 \\
C-PER    &                     & 9.34  & 0.7780 &                     & 11.17 & 0.9305 &                     & 8.61  & 0.8606 \\
C-PER+at &                     & 9.38  & 0.7816 &                     & 11.74 & 0.9784 &                     & 8.68  & 0.8681 \\
\Xhline{4\arrayrulewidth}
\end{tabular}
\caption{Number of steps of bound calculation for the optimal $\epsilon$ in Fast-Lin (T\textsubscript{Lin}) and PEC (T\textsubscript{PEC}) for ReLU networks under $l_\infty$ attacks. Note that T\textsubscript{Lin} is a constant for different models given the original interval $[\underline{\epsilon}, \bar{\epsilon}]$.} \label{tbl:search}
\vspace{-0.2cm}
\end{table*}

We show the results on $l_\infty$ in Table~\ref{tbl:search} and defer the $l_2$ results in Table~\ref{tbl:app_search_1} of Appendix~\ref{subsec:app_optimal_eps}.
For $l_\infty$ cases, the original interval $[\underline{\epsilon}, \bar{\epsilon}]$ is $[0, 0.4]$ for MNIST and $[0, 0.1]$ for CIFAR10.
Note that, because PEC has almost no computational overhead compared with Fast-Lin and CROWN,\footnote{We run one-iteration PEC and CROWN to certify CIFAR10-CNN models by C-PER+at for 10 times. To process the entire test set on a single GPU machine, in $l_\infty$ cases, the mean and standard deviation of run time is $217.51 \pm 1.95$ seconds for CROWN and $219.16 \pm 3.23$ seconds for PEC; in $l_2$ cases, it is $236.95 \pm 1.64$ for CROWN and $239.41 \pm 1.92$ for PEC. Therefore the difference can be ignored.} the number of iterations reflects the running time to obtain the optimal certified bounds.
Altogether, our results show that PEC can save approximately $25\%$ of the running time for FC1 models and $10\%$ of the running time for CNN models.

\begin{figure}
\centering
\includegraphics[scale = 0.45]{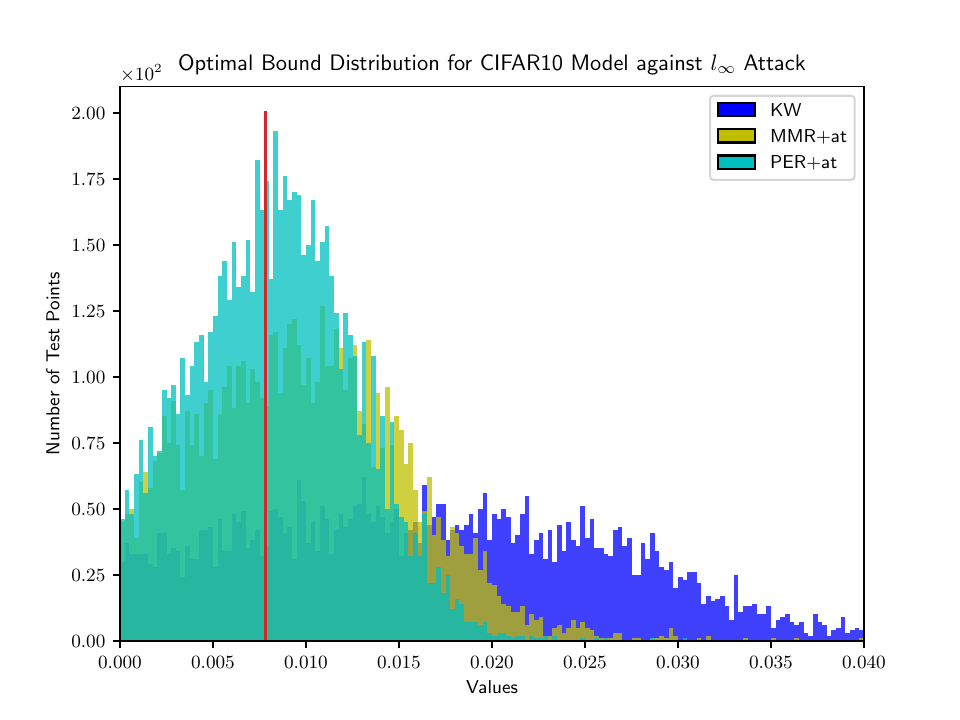}
\caption{Distribution of optimal certified bounds of CIFAR10 models trained against $l_\infty$ attacks. The target bound ($2 / 255$) is indicated by a red vertical line.} \label{fig:optimal_eps_dist}
\vspace{-0.2cm}
\end{figure}

Figure~\ref{fig:optimal_eps_dist} shows the distribution of the optimal certified bounds for CIFAR10 models against $l_\infty$ attacks obtained by KW, MMR+at and C-PER+at on the test set.
The results on $l_2$ attacks are shown in Figure~\ref{fig:optimal_eps_dist_app} of Appendix~\ref{subsec:app_opt_bounds}.
We use vertical red lines to represent the target bounds ($2/255$ in the $l_\infty$ case and $0.1$ in the $l_2$ case), so the area on the right of this line represents the certified robust accuracy.
Compared with KW, the mass of C-PER+at is more concentrated on a narrower range on the right of the red line.
This evidences that there are significantly fewer points that have unnecessarily large certified bounds for the C-PER+at model than for the KW one.
This is because PER+at encourages robustness via a hinge-loss term.
When $\tilde{d}_{ic} \geq \alpha$, the regularizer in Equation (\ref{eq:per_def}) is a constant zero and does not contribute to the parameter gradient.
However, KW first estimates the bound of the worst case output logits and calculates the softmax cross-entropy loss on that.
Under this training objective function, each data point is encouraged to make the lower bound of the true label's output logit bigger and the upper bound of false ones smaller, even if the current model is sufficiently robust at this point.
This phenomenon also helps to explain why KW tends to over-regulate the model while our methods do not.

\change{
We have also tried to replace the cross-entropy loss with the hinge loss in the objective of KW, but observed this not to lead to any improvement over the original KW. This is because KW directly minimizes the gap between the logits of the true and false label, but the logits' magnitude for different instances differs, which makes it difficult or even impossible to set a unified threshold in the hinge loss.
By contrast, in PER, we apply the hinge loss to the certified bound directly, which is normalized, easier to interpret and thus makes it much easier to set the threshold in the hinge loss.
In practice, the value of $\alpha$ in Equation~\ref{eq:per_def} is set $1.5$ times the target adversarial budget.
}


\section{Discussion} \label{sec:discussion}

\begin{table}[h]
\normalsize
\centering
\begin{tabular}{p{3.6cm}p{2.7cm}<{\raggedleft\arraybackslash}}
\Xhline{4\arrayrulewidth}
Methods & Complexity\\
\hline
PGD              & $\bigO(N n^2)$    \\
Fast-Lin / CROWN & $\bigO(N^2 n^3)$  \\
KW               & $\bigO(N^2 n^3)$  \\
MMR / MMR+at     & $\bigO(N n^2 m)$  \\
IBP              & $\bigO(N n^2)$    \\
C-IBP            & $\bigO(N n^3)$    \\
I-PER / I-PER+at & $\bigO(N n^2 m)$  \\
C-PER / C-PER+at & $\bigO(N^2 n^3)$  \\
\Xhline{4\arrayrulewidth}
\end{tabular}
\caption{\change{Complexity of different methods on an $N$-layer neural network model with $k$-dimensional output and $m$-dimensional input. Each hidden layer has $n$ neurons.}}
\label{tab:complexity}
\vspace{-0.2cm}
\end{table}

Let us consider an $N$-layer neural network model with $k$-dimensional output and $m$-dimensional input.
For simplicity, let each hidden layer have $n$ neurons and usually $n \gg \max\{k, m\}$ is satisfied.
In this context, the FLOP complexity of PGD with $h$ iterations is $\bigO(N n^2 h) \sim \bigO(N n^2)$, because typically $h \ll \min\{m, n\}$.
\change{
Among the methods that train provably robust networks, the linearization algorithm based on Fast-Lin / CROWN needs $\bigO(N^2 n^3)$ FLOPS to obtain the linear bounds of the output logits.
However, the complexity can be reduced to $\bigO(N n^2 m)$ at the cost of bound tightness when we use the IBP-inspired algorithm in Appendix~\ref{subsec:app_ibp}.
Note that the IBP-inspired algorithm also calculates the linear bound of the output logits and is thus different from IBP \cite{gowal2018effectiveness}, whose complexity is $\bigO(Nn^2)$, i.e., the same as a forward propagation.
MIP is a complete certifier based on mixed integer programming.
It solves an NP-hard problem and its complexity is super-polynomial in general.
To update the model parameters, KW needs a back-propagation which costs $\bigO(N n^2)$ FLOPs.
Therefore the complexity of KW is also $\bigO(N^2 n^3)$, with the model linearization dominating the complexity.
In CROWN-IBP, the bounds of all intermediate layers are estimated by IBP, which costs $\bigO(N n^2)$ FLOPs.
The last layer's bound is then estimated in the same way as CROWN, which costs $\bigO(N n^3)$ FLOPs, dominating the complexity of CROWN-IBP.
For MMR, the complexity to calculate the expression of the input's linear region is $\bigO(N n ^2 m)$.
MMR then calculates the distances between the input and $\bigO(N n)$ hyper-planes, costing $\bigO(N n m)$.
Altogether, the complexity of MMR is $\bigO(N n^2 m)$.
MMR+at has the same complexity as MMR, because the overhead of adversarial training can be ignored.

Among our methods, the complexity of the CROWN-style model linearization in C-PER is $\bigO(N^2 n^3)$.
Like MMR and KW, the overhead of distance calculation and back-propagation can be ignored.
Similarly, the complexity of I-PER is dominated by the IBP-inspired model linearization, which is $\bigO(N n^2 m)$.
Note that C-PER has the same complexity as Fast-Lin, CROWN and KW, and the complexity of I-PER is smaller than that of CROWN-IBP because $m < n$.
C-PER+at and I-PER+at have the same complexity as C-PER and I-PER, respectively, since the overhead of adversarial training is negligible.
Table~\ref{tab:complexity} summarizes the complexity of all methods.
}

No matter which linearization method we use, the bounds of the output logits inevitably become looser for deeper networks, which can be a problem for large models.
Furthermore, the linear approximation implicitly favors the $l_\infty$ norm over other $l_p$ norms because the intermediate bounds are calculated in an elementwise manner \cite{liu2019certifying}.
As a result, our method performs better in $l_\infty$ cases than in $l_2$ cases.
Designing a training algorithm with scalable and tight certified robustness is highly non-trivial and worth further exploration.


\section{Conclusion} \label{sec:conclusion}

In this paper, we have studied the robustness of neural networks from a geometric perspective.
In our framework, linear bounds are estimated for the model's output under an adversarial budget.
\change{
Then, the polyhedral envelope resulting from the linear bounds allows us to obtain quantitative robustness guarantees.
Our certification method can give non-trivial robustness guarantees to more data points than existing methods and thus speed up the search for the optimal adversarial budget's size.
Furthermore, we have shown that our certified bounds can be turned into a geometry-inspired regularization scheme that enables training provably robust models.
Compared with existing methods, our framework can be applied to neural networks with general activation functions.
In addition to better performance, it can achieve provable robustness at very little loss in clean accuracy.
}

\section{Acknowledgement}

We thankfully acknowledge the support of the Hasler Foundation (Grant No. 16076) for this work.

\appendices

\section{Model Linearization} \label{sec:app_linearize}

\subsection{Linearization of Activation Functions} \label{subsec:app_linearize_activation}

In this section, we discuss the choice of $d$, $l$ and $h$ in the linear approximation $dx + l \leq \sigma(x) \leq dx + h$ for activation function $\sigma$ when $x \in [\underline{x}, \bar{x}]$.
The method used here is slightly different from that of~\cite{zhang2018efficient}.
First, the slope of the linear upper and lower bound is the same, because this can save up to $3/4$ memory when calculating the slope of the linear bound.
Second, all coefficients need to have an analytical form because we need to calculate the gradient based on them during training.
Note that~\cite{zhang2018efficient} use binary search to obtain the optimal $d$, $l$, $h$ for general activation functions.

\subsubsection{ReLU}

\begin{figure}[h]
\centering
\subfigure[$\underline{x} \leq \bar{x} \leq 0$]{\includegraphics[scale = 0.165]{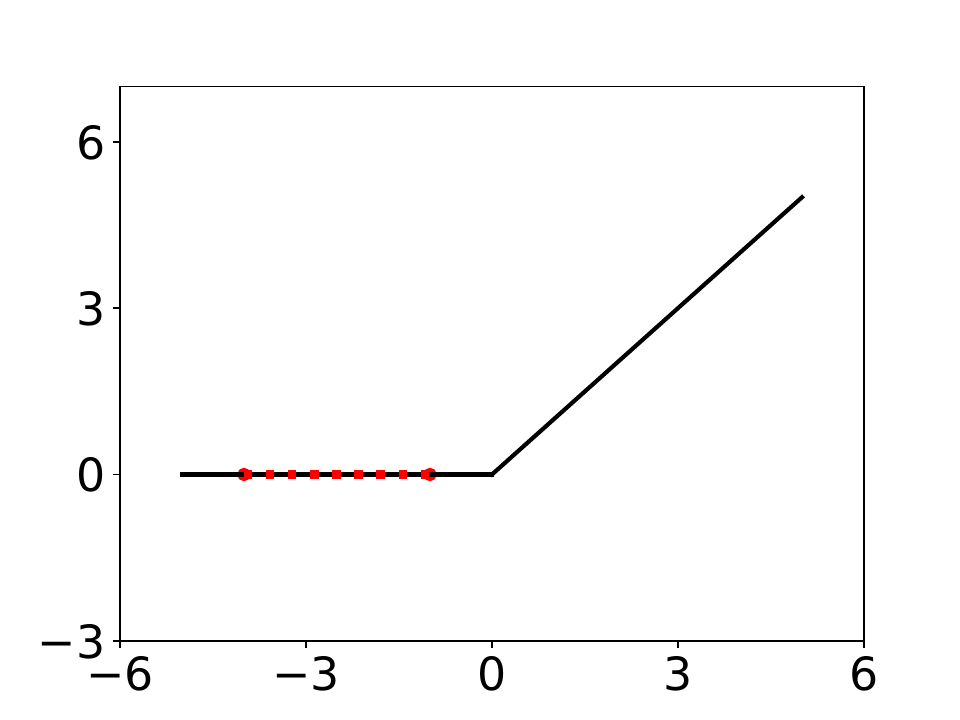}}
\subfigure[$\underline{x} < 0 < \bar{x}$]{\includegraphics[scale = 0.165]{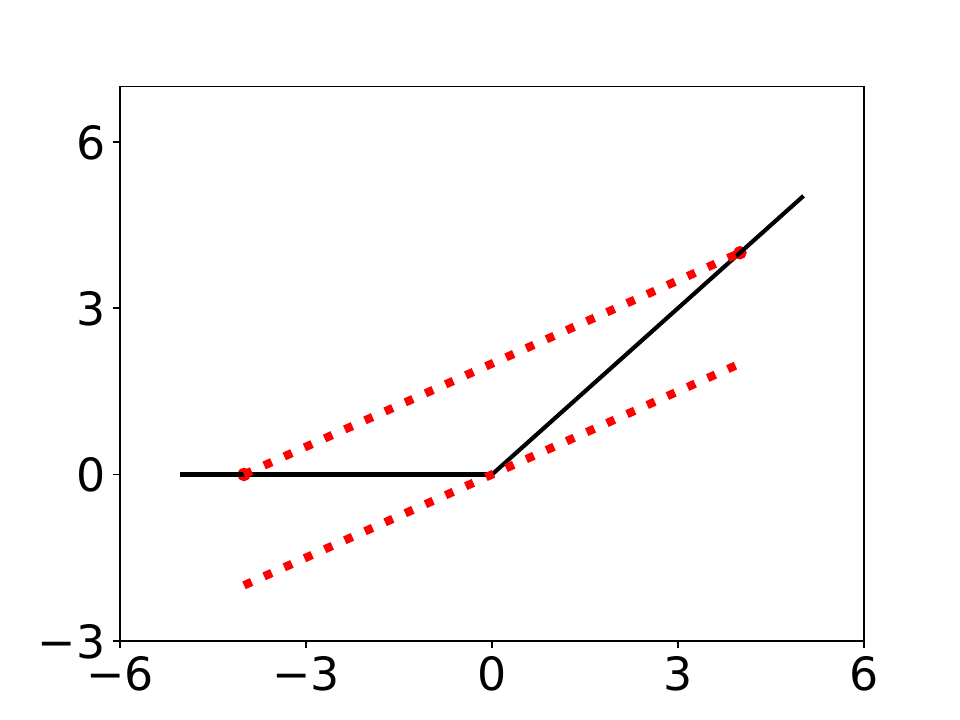}}
\subfigure[$0 \leq \underline{x} \leq \bar{x}$]{\includegraphics[scale = 0.165]{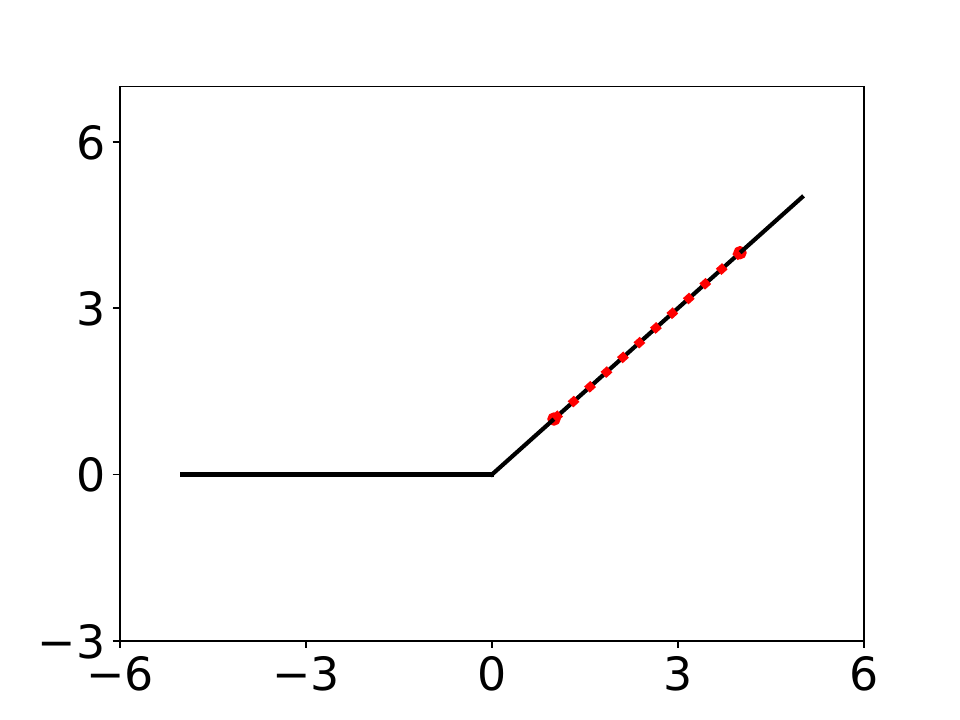}}
\caption{Linearization of the ReLU function in all scenarios.} \label{fig:relu_linearize}
\vspace{-0.2cm}
\end{figure}

As Figure \ref{fig:relu_linearize} shows, the linear approximation for ReLU $\sigma(x) = \max(0, x)$, which is convex, is:

\begin{equation}
\begin{aligned}
d &= \left\{
\begin{aligned}
&0 & \underline{x} \leq \bar{x} \leq 0 \\
&\frac{\bar{x}}{\bar{x} - \underline{x}} & \underline{x} < 0 < \bar{x} \\
&1 & 0 \leq \underline{x} \leq \bar{x}
\end{aligned}
\right.&, \\
l &= 0\ \forall x \in \R, \\
h &= \left\{
\begin{aligned}
&0 & \underline{x} \leq \bar{x} \leq 0 \\
&- \frac{\underline{x}\bar{x}}{\bar{x} - \underline{x}} & \underline{x} < 0 < \bar{x} \\
&0 & 0 \leq \underline{x} \leq \bar{x}
\end{aligned}
\right. &.
\end{aligned}
\end{equation}

\subsubsection{Sigmoid, Tanh}

\begin{figure}
\centering
\subfigure[$\underline{x} \leq \bar{x} \leq 0$]{\includegraphics[scale = 0.165]{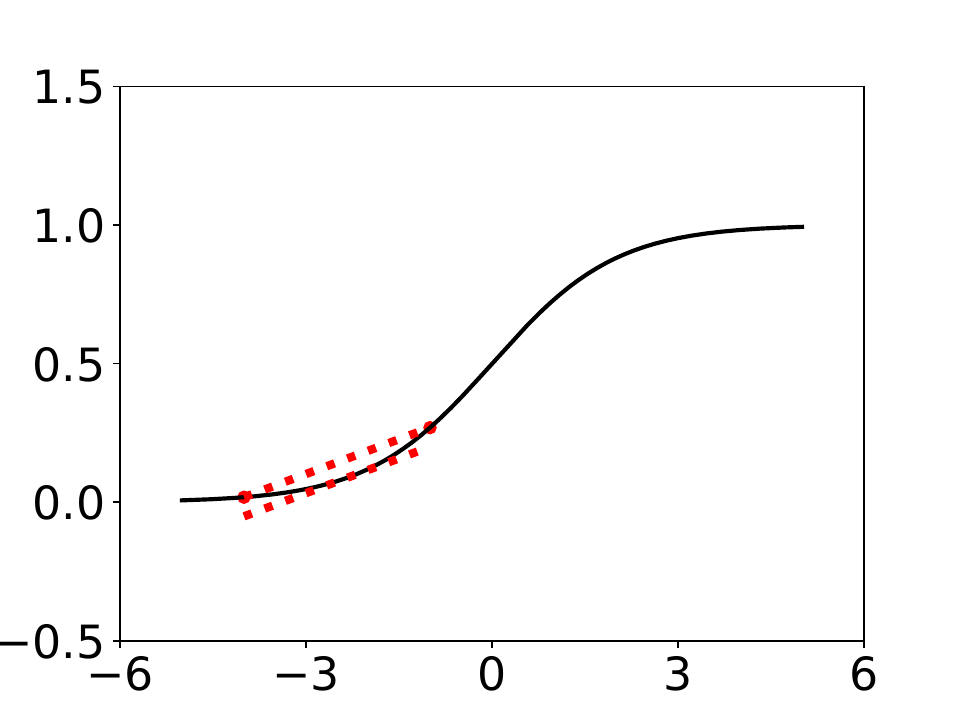}}
\subfigure[$\underline{x} < 0 < \bar{x}$]{\includegraphics[scale = 0.165]{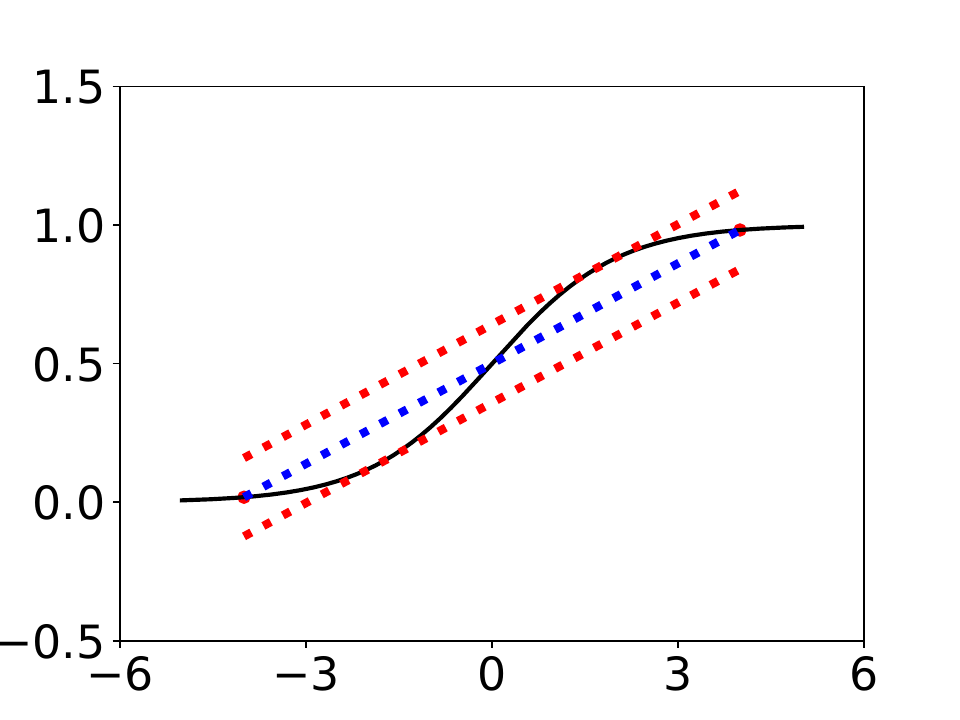}}
\subfigure[$0 \leq \underline{x} \leq \bar{x}$]{\includegraphics[scale = 0.165]{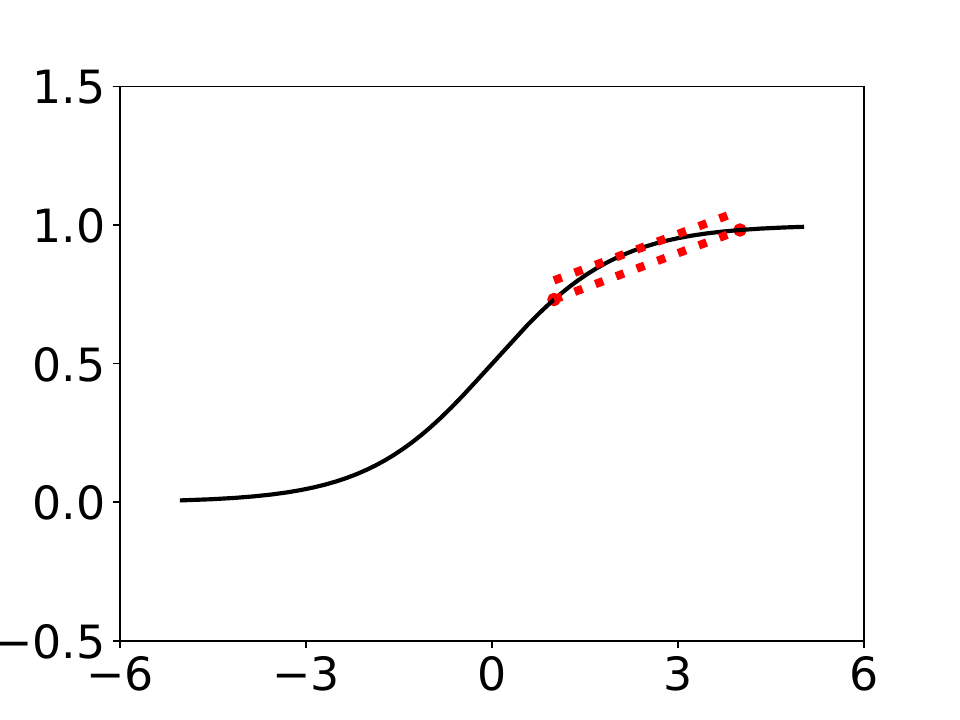}}
\caption{Linearization of the sigmoid function in all scenarios.} \label{fig:sigd_linearize}
\vspace{-0.2cm}
\end{figure}

Unlike the ReLU function, the sigmoid function $\sigma(x) = \frac{1}{1 + e^{-x}}$ and tanh function $\sigma(x) = \frac{e^{2x} - 1}{e^{2x} + 1}$ are not convex.
However, these two functions are convex when $x < 0$ and concave when $x > 0$ (left and right sub-figures of Figure~\ref{fig:sigd_linearize}).
Therefore, when $\underline{x} \leq \bar{x} \leq 0$ or $0 \leq \underline{x} \leq \bar{x}$, we can easily obtain a tight linear approximation.
When $\underline{x} \leq 0 \leq \bar{x}$, we do not use the binary research to obtain a tight linear approximation as in~\cite{zhang2018efficient}, because the results would not have an analytical form in this way.
Instead, we first calculate the slope between the two ends, i.e., $d = \frac{\sigma(\bar{x}) - \sigma(\underline{x})}{\bar{x} - \underline{x}}$.
Then, we bound the function by two tangent lines of the same slope as $d$ (middle sub-figure of Figure~\ref{fig:sigd_linearize}).

For sigmoid and tanh, we can calculate the coefficients of the linear approximation as
\begin{equation}
\begin{aligned}
d &= \frac{\sigma(\bar{x}) - \sigma(\underline{x})}{\bar{x} - \underline{x}}., \\
l &= \left\{
\begin{aligned}
& \sigma(t_1) - t_1 d & \underline{x} < 0 \\
& \frac{\bar{x} \sigma(\underline{x}) - \underline{x} \sigma(\bar{x})}{\bar{x} - \underline{x}} & 0 \leq \underline{x} \leq \bar{x}
\end{aligned}
\right.,\\
h &= \left\{
\begin{aligned}
& \frac{\bar{x} \sigma(\underline{x}) - \underline{x} \sigma(\bar{x})}{\bar{x} - \underline{x}} & \underline{x} \leq \bar{x} \leq 0 \\
& \sigma(t_2) - t_2 d & 0 < \bar{x} \\
\end{aligned}
\right. .
\end{aligned}
\end{equation}

The coefficients $t_1 < 0 < t_2$ are the position of tangent points on both sides of the origin.
The definitions of $t_1$ and $t_2$ for different activation functions are provided in Table \ref{tbl:t}.

\begin{table}[h]
\centering
\small
\begin{tabular}{ccc}
\Xhline{4\arrayrulewidth}
$\sigma$ & Sigmoid & Tanh\\
\hline
\\[-0.2cm]
$t_1$ & $-\log{\frac{-(2d - 1) + \sqrt{1 - 4d}}{2d}}$ & $\frac{1}{2}\log{\frac{-(d - 2) - 2\sqrt{1 - d}}{d}}$ \\[0.2cm]
$t_2$ & $-\log{\frac{-(2d - 1) - \sqrt{1 - 4d}}{2d}}$ & $\frac{1}{2}\log{\frac{-(d - 2) + 2\sqrt{1 - d}}{d}}$ \\[0.1cm]
\Xhline{4\arrayrulewidth}
\end{tabular}
\vspace{0.1cm}
\caption{Definition of $t_1$ and $t_2$ for different activation functions.} \label{tbl:t}
\vspace{-0.4cm}
\end{table}

\subsection{CROWN-style Bounds} \label{subsec:app_fastlin_crown}

Based on the linear approximation of activation functions in Section~\ref{subsec:app_linearize_activation}, we have $\D^{(i)}\z'^{(i)} + \low^{(i)} \leq \sigma(\z'^{(i)}) \leq \D^{(i)}\z'^{(i)} + \up^{(i)}$ where $\D^{(i)}$ is a diagonal matrix and $\low^{(i)}$, $\up^{(i)}$ are vectors.
We can rewrite this formulation as follows:

\begin{equation}
\begin{aligned}
&\exists \D^{(i)}, \low^{(i)}, \up^{(i)}:\ \forall \z'^{(i)} \in [\underline{\z}^{(i)},\ \bar{\z}^{(i)}],\\
\mathrm{then}\ &\exists \m^{(i)} \in [\low^{(i)}, \up^{(i)}]\ s.t. \sigma(\z'^{(i)}) = \D^{(i)}\z'^{(i)} + \m^{(i)}\;.
\end{aligned} \label{eq:approx}
\end{equation}

We plug (\ref{eq:approx}) into~(\ref{eq:model}), and the expression of $\z'^{(i)}$ can be rewritten as

\begin{equation}
\scriptsize
\begin{aligned}
&\z'^{(i)} = \W^{(i - 1)}(\sigma( \W^{(i - 2)} (... \sigma(\W^{(1)}\hat{\z}'^{(1)} + \bias^{(1)}) ...)+ \bias^{(i - 2)})) + \bias^{(i - 1)} \\
&= \W^{(i - 1)}(\D^{(i - 1)}( \W^{(i - 2)} (... \D^{(2)}(\W^{(1)}\x' + \bias^{(1)}) + \m^{(2)} ...) \\ &+ \bias^{(i - 2)}) + \m^{(i - 1)}) + \bias^{(i - 1)} \\
&= \left(\prod_{k = 1}^{i - 1} \W^{(k)}\D^{(k)} \right) \W^{(1)}\x' + \sum_{j = 1}^{i - 1}\left(\prod_{k = j + 1}^{i - 1} \W^{(k)}\D^{(k)}\right)\bias^{(j)} \\ &+ \sum_{j = 2}^{i - 1}\left(\prod_{k = j + 1}^{i - 1} \W^{(k)}\D^{(k)}\right)\W^{(j)}\m^{(j)}\;.
\end{aligned} \label{eq:approx_x}
\end{equation}

This is a linear function w.r.t. $\x'$ and $\{\m^{(j)}\}_{j = 2}^{i - 1}$.
Once given the perturbation budget $\Set_{\epsilon}^{(p)}(\x)$ and the bounds of $\{\m^{(j)}\}_{j = 2}^{i - 1}$, we can calculate the slope and the bias term in the linear bound of $\z'^{(i)}$ in Equation~(\ref{eq:approx_x}).
This process can be repeated until we obtain the bound of the output logits in Equation~(\ref{eq:linearize}).
The derivation here is the same as in~\cite{liu2019certifying, weng2018towards}, we encourage interested readers to check these works for details.

\subsection{IBP-inspired Bounds} \label{subsec:app_ibp}

Interval Bound Propagation (IBP), introduced in~\cite{gowal2018effectiveness}, is a simple and scalable method to estimate the bounds of each layer in neural networks.
IBP is much faster than the algorithm introduced in Appendix~\ref{subsec:app_fastlin_crown} because the bounds of any intermediate layer are calculated only based on the information of its immediate previous layer.
Therefore, the bounds are propagated just like inference in network models, which costs only $\bigO(N)$ matrix-vector multiplications for an $N$-layer network defined in (\ref{eq:model}).

In our work, we need linear bounds of the output logits in addition to general numeric bounds, so the linearization of activation functions defined in (\ref{eq:approx}) is necessary.
We define linear bounds $\U^{(i)}\x' + \p^{(i)} \leq \z'^{(i)} \leq \U^{(i)}\x' + \q^{(i)}$, $\widehat{\U}^{(i)}\x' + \hat{\p}^{(i)} \leq \hat{\z}'^{(i)} \leq \widehat{\U}^{(i)}\x' + \hat{\q}^{(i)}$.
We use the same slope as in Section~\ref{subsec:app_linearize_activation} to linearize the activation functions, so the slopes of both bounds are the same.
Plugging~(\ref{eq:approx}) into this formulation, we have

\begin{equation}
\begin{aligned}
\widehat{\U}^{(i)} &= \D^{(i)}\U^{(i)},\\
\hat{\p}^{(i)} &= \D^{(i)}\p^{(i)}+ \low^{(i)},\\
\hat{\q}^{(i)} &= \D^{(i)}\q^{(i)} + \up^{(i)}\;.
\end{aligned} \label{eq:iter1}
\end{equation}

Here, we assume that the activation functions are monotonically increasing, so the elements in $\D^{(i)}$ are non-negative.
Similarly, by comparing the linear bounds of $\hat{\z}'^{(i)}$ and $\z'^{(i + 1)}$, we have

\begin{equation}
\begin{aligned}
\U^{(i + 1)} &= \W^{(i)}\widehat{\U}^{(i)},\\
\p^{(i + 1)} &= \W^{(i)}_{+}\hat{\p}^{(i)} + \W^{(i)}_{-}\hat{\q}^{(i)} + \bias^{(i)},\\
\q^{(i + 1)} &= \W^{(i)}_{+}\hat{\q}^{(i)} + \W^{(i)}_{-}\hat{\p}^{(i)} + \bias^{(i)}\;.
\end{aligned} \label{eq:iter2}
\end{equation}

By definition, we have $\widehat{\U}^{(1)} = \mathbf{I}$ and $\hat{\p}^{(1)} = \hat{\q}^{(1)} = \mathbf{0}$.
Applying Equation (\ref{eq:iter1}) and (\ref{eq:iter2}) iteratively allows us to obtain the values of the coefficients $\U^{(N)}$, $\V^{(N)}$, $\p^{(N)}$ and $\q^{(N)}$ in Equation (\ref{eq:linearize}).

\section{Algorithms}

\subsection{Algorithms for Searching the Optimal Value of $\epsilon$} \label{subsec:app_alg_search}

The pseudo code for finding the optimal $\epsilon$ is provided as Algorithm~\ref{alg:app_search} below.
$\mathcal{M}$, $\x$, $\epsilon_{\Delta}$, $\underline{\epsilon}$, $\bar{\epsilon}$ represent the classification model, the input point, the precision requirement, the predefined estimate of the lower bound and of the upper bound, respectively.
Typically, $\underline{\epsilon}$ is set to $0$ and $\bar{\epsilon}$ is set to a large value corresponding to a perceptible the image perturbation.
$f$ is a function mapping a model, an input point and a value of $\epsilon$ to a certified bound.
$f$ is a generalized form of the Fast-Lin, CROWN and PEC algorithms.

During the search for the optimal $\epsilon$, the lower bound is updated by the current certified bound, while the upper bound is updated when the current certified bound is smaller than the choice of $\epsilon$.
In Fast-Lin and CROWN, we update either the lower or the upper bound in one iteration since the certified bound is either $0$ or the current choice of $\epsilon$.
However, it is possible for PEC to update both the lower and the upper bounds in one iteration, and this leads to a faster convergence of $\epsilon$.

\begin{algorithm}
\begin{algorithmic}
\STATE \textbf{Input:} $\x$, $\underline{\epsilon}$, $\bar{\epsilon}$, $\epsilon_{\Delta}$, $f$, $\mathcal{M}$
\STATE Set the bounds of $\epsilon$: $\epsilon_{up} = \bar{\epsilon}$, $\epsilon_{low} = \underline{\epsilon}$
\WHILE {$\epsilon_{up} - \epsilon_{low} > \epsilon_{\Delta}$}
    \STATE $\epsilon_{try}$ = $\frac{1}{2}(\epsilon_{low} + \epsilon_{up})$
    \STATE $\epsilon_{cert} = f(\mathcal{M}, \x, \epsilon_{try})$
    \STATE Update lower bound: $\epsilon_{low} = \max\{\epsilon_{low}, \epsilon_{cert}\}$
    \IF {$\epsilon_{try} > \epsilon_{cert}$}
        \STATE Update upper bound: $\epsilon_{up} = \epsilon_{try}$
    \ENDIF
\ENDWHILE
\STATE \textbf{Output:} $\frac{1}{2}(\epsilon_{low} + \epsilon_{up})$
\end{algorithmic}
\caption{Search for optimal value of $\epsilon$} \label{alg:app_search}
\end{algorithm}

\section{Proofs} \label{sec:app_proof}

\subsection{Proof of Lemma \ref{theory:bounds}} \label{subsec:proof_bound_thm}

\begin{proof}
Let $\x' = \x + \Delta$ be a point that breaks condition (\ref{eq:sufficient_condition}). Then,

\begin{equation}
\begin{aligned}
& && \U_i(\x + \Delta) + \p_i  &\ <\ & 0 \\
\iff & && \U_i\Delta &\ <\ & - \U_i\x - \p_i \\
\Longrightarrow & && -\|\U_i\|_q\|\Delta\|_p &\ <\ & - \U_i\x - \p_i \\
\iff & && \|\Delta\|_p &\ >\ & \frac{\U_i\x + \p_i}{\|\U_i\|_q}
\end{aligned} \label{eq:proof_thm1}
\end{equation}

The $\Longrightarrow$ comes from H\"older's inequality.
(\ref{eq:proof_thm1}) indicates that a perturbation of $l_p$ norm over $d_{ic} = \max\left\{0, \frac{\U_i\x + \p_i}{\|\U_i\|_q}\right\}$ is needed to break the sufficient condition of $\z'^{(N)}_c - \z'^{(N)}_i \geq 0$.
Based on the assumption of adversarial budget $\Set^{(p)}_\epsilon(\x)$ when linearizing the model, the $l_p$ norm of a perturbation to produce an adversarial example is at least $\min\left\{\epsilon, d_c\right\}$.
\end{proof}

\subsection{Proof of Theorem~\ref{coro:optimal}} \label{subsec:proof_constrained_bounds}

\begin{proof}

We use the primal-dual method to solve the optimization problem (\ref{eq:bounded_constrained_problem}), which is a convex optimization problem with linear constraints.

It is clear that there exists an image inside the allowable pixel space for which the model predicts the wrong label.
That is, the constrained problem (\ref{eq:bounded_constrained_problem}) is strictly feasible:

\begin{equation}
\begin{aligned}
\exists \Delta\ s.t.\ \ai\Delta + b < 0, \Delta^{(min)} < \Delta < \Delta^{(max)}\;.
\end{aligned}
\end{equation}

Thus, this convex optimization problem satisfies \textit{Slater's Condition}, i.e., strong duality holds. We then rewrite the primal problem as

\begin{equation}
\begin{aligned}
& \min_{\Delta^{(min)} \leq \Delta \leq \Delta^{(max)}} \|\Delta\|^p_p & \\
& ~~~~~s.t.\ \ \ai\Delta + b \leq 0
\end{aligned} \label{eq:constrained_primal}
\end{equation}

We minimize $\|\Delta\|^p_p$ instead of directly $\|\Delta\|_p$ in order to decouple all elements in vector $\Delta$.
In addition, we consider $\Delta^{(min)} \leq \Delta \leq \Delta^{(max)}$ as the domain of $\Delta$ instead of constraints for simplicity.
We write the dual problem of (\ref{eq:constrained_primal}) by introducing a coefficient of relaxation $\lambda \in \mathbb{R}_{+}$:

\begin{equation}
\begin{aligned}
\max_{\lambda \geq 0} \min_{\Delta^{(min)} \leq \Delta \leq \Delta^{(max)}} g(\Delta, \lambda) \eqdef \|\Delta\|^p_p + \lambda (\ai \Delta + b)
\end{aligned}
\end{equation}

To solve the inner minimization problem, we set the gradient $\frac{\partial g(\Delta, \lambda)}{\partial \Delta_i} = \text{sign}(\Delta_i) p |\Delta_i|^{p - 1} + \lambda \ai_i$
 to zero and obtain $\Delta_i = - \text{sign}(\ai_i) \left|\frac{\lambda \ai_i}{p}\right|^{\frac{1}{p - 1}}$.
Based on the convexity of function $g(\Delta, \lambda)$ w.r.t. $\Delta$, we can obtain the optimal $\tilde{\Delta}_i$ in the domain:

\begin{equation}
\small
\begin{aligned}
\tilde{\Delta}_i = \text{clip}\left(-\text{sign}(\ai_i) \left|\frac{\lambda \ai_i}{p}\right|^{\frac{1}{p - 1}}, \text{min} = \Delta^{(min)}_i, \text{max} = \Delta^{(max)}_i\right)\;.
\end{aligned} \label{eq:solution}
\end{equation}

Based on strong duality, we can say that the optimal $\tilde{\Delta}$ is chosen by setting a proper value of $\lambda$.
Fortunately, $\|\tilde{\Delta}\|_p$ increases monotonically with $\lambda$, so the smallest $\lambda$ corresponds to the optimum.

As we can see, the expression of $\widehat{\Delta}$ in (\ref{eq:holder_optimal}) is consistent with $\tilde{\Delta}_i$ in (\ref{eq:solution}) if $\lambda$ is set properly.\footnote{The power term $\frac{q}{p} = \frac{1}{p - 1}$ when $\frac{1}{p} + \frac{1}{q} = 1$}
The greedy algorithm in Algorithm \ref{alg:greedy} describes the process of gradually increasing $\lambda$ to find the smallest value satisfying the constraint $\ai\Delta + b \leq 0$.
With the increase of $\lambda$, the elements in vector $\Delta$ remain unchanged when they reach either $\Delta^{(min)}$ or $\Delta^{(max)}$, so we keep such elements fixed and optimize the others.

\end{proof}

\section{Additional Experiments}

\subsection{Details of the Experiments} \label{subsec:app_experiment_details}

\subsubsection{Model Architecture}

The FC1 and CNN networks used in this paper are identical to the ones used in \cite{croce2018provable}.
The FC1 network is a fully-connected network with one hidden layer of 1024 neurons.
The CNN network has two convolutional layers and one additional hidden layer before the output layer.
Both convolutional layers have a kernel size of 4, a stride of 2 and a padding of 1 on both sides, so the height and width of the feature maps are halved after each convolutional layer.
The first convolutional layer has 32 channels while the second one has 16.
The hidden layer following them has 100 neurons.

\subsubsection{Hyper-parameter Settings}

In all experiments, we use the Adam optimizer~\cite{kingma2014adam} with an initial learning rate of $10^{-3}$ and train all models for 100 epochs with a mini-batch of 100 instances.
For CNN models, we decrease the learning rate to $10^{-4}$ for the last 10 epochs.
When we train CNN models on MNIST, we only calculate the polyhedral envelope of 20 instances subsampled from each mini-batch.
When we train CNN models on CIFAR10, this subsampling number is $10$.
These settings make our algorithm possible to be trained on a GPU with 12 GB memory.
For PER and PER+at, the value of $T$ is always $4$.
We search in the logarithmic scale for the value of $\gamma$ and in the linear scale for the value of $\alpha$.
For $\epsilon$, we ensure that its values in the end of training are close to the ones used in the adversarial budget $\Set^{(p)}_\epsilon(\x)$.
We compare constant values with an exponential growth scheme for $\epsilon$ but always use constant values for $\alpha$ and $\gamma$.
The optimal values we found for different settings are provided in Table~\ref{tbl:hyperparams}.

\begin{table}[h]
\centering
\begin{tabular}{p{1.2cm}p{0.8cm}<{\centering}p{2.4cm}<{\centering}p{1.8cm}<{\centering}}
\Xhline{4\arrayrulewidth}
Task & $\alpha$ & $\epsilon$ & $\gamma$ \\
\hline
MNIST &\multirow{2}{*}{$0.15$} & initial value $0.0064$ & \multirow{2}{*}{$0.1$} \\
FC1, $l_\infty$ & & $\times 2$ every 20 epochs & \\[0.2cm]
MNIST & \multirow{2}{*}{$0.15$} & \multirow{2}{*}{$0.1$} & PER: $0.3$ \\
CNN, $l_\infty$ & & & PER+at: $0.03$ \\[0.2cm]
CIFAR10 & \multirow{2}{*}{$0.1$} & \multirow{2}{*}{$0.008$} & PER: $0.0003$ \\
CNN, $l_\infty$ & & & PER+at: $0.001$ \\[0.2cm]
MNIST & \multirow{2}{*}{$0.45$} & initial value $0.02$ & \multirow{2}{*}{$1.0$} \\
FC1, $l_2$ & & $\times 2$ every 20 epochs & \\[0.2cm]
MNIST & \multirow{2}{*}{$0.45$} & \multirow{2}{*}{$0.3$} & \multirow{2}{*}{$1.0$} \\
CNN, $l_2$ & & & \\[0.2cm]
CIFAR10 & \multirow{2}{*}{$0.15$} & \multirow{2}{*}{$0.1$} & PER: $0.3$ \\
CNN, $l_2$ & & & PER+at: $1.0$ \\[0.1cm]
\Xhline{4\arrayrulewidth}
\end{tabular}
\caption{Values of $\alpha$, $\epsilon$ and $\gamma$ for different experiments.} \label{tbl:hyperparams}
\vspace{-0.4cm}
\end{table}

\subsection{Additional Experimental Results} \label{subsec:app_extra_experiments}

\subsubsection{$l_2$ Robustness on ReLU Networks} \label{subsubsec:app_relu_l2}

The results of 11 training methods and 7 evaluation metrics on $l_2$ robustness are provided in Table~\ref{tbl:main_l2}.
In all three cases studied, our proposed methods, either PER or PER+at, achieves the best performance.

\begin{table*}[h]
\small
\centering
\begin{tabular}{p{2.2cm}p{1.3cm}<{\raggedleft\arraybackslash}p{1.3cm}<{\raggedleft\arraybackslash}p{1.3cm}<{\raggedleft\arraybackslash}p{1.3cm}<{\raggedleft\arraybackslash}p{1.3cm}<{\raggedleft\arraybackslash}p{1.3cm}<{\raggedleft\arraybackslash}p{1.3cm}<{\raggedleft\arraybackslash}}
\Xhline{4\arrayrulewidth}
Methods & CTE ~~~ (\%) & PGD ~~~ (\%) & CRE Lin (\%) & CRE IBP (\%) & ACB Lin & ACB IBP & ACB PEC \\
\hline
\\[-0.2cm]
& \multicolumn{7}{c}{\textbf{MNIST - FC1, ReLU, $l_2$, $\epsilon = 0.3$}} \\[0.1cm]
plain               & 1.99     & 9.81     & 40.97    & 99.30    & 0.1771     & 0.0021     & 0.2300     \\
at                  & 1.35     & 2.99     & 14.85    & 99.23    & 0.2555     & 0.0023     & 0.2684     \\
KW                  & 1.23     & 2.70     & \two{4.91} & 41.55    & \two{0.2853} & 0.1754     & \two0.2892 \\
IBP                 & 1.36     & 2.90     & 6.87     & \two{9.01} & 0.2794     & \two{0.2730} & 0.2876     \\
C-IBP               & 1.26     & 2.80     & 6.36     & \one{8.73} & 0.2809     & \one{0.2738} & 0.2884     \\
MMR                 & 2.40     & 5.88     & 7.76     & 99.55    & 0.2767     & 0.0013     & 0.2845     \\
MMR+at              & 1.77     & 3.76     & 5.68     & 99.86    & 0.2830     & 0.0004     & 0.2880     \\
\textbf{C-PER}      & 1.26     & 2.44     & 5.35     & 59.17    & 0.2840     & 0.1225     & 0.2888     \\
\textbf{C-PER+at}   & \one{0.67} & \one{1.40} & \one{4.84} & 64.79    & \one{0.2855} & 0.1056     & \one{0.2910} \\
\textbf{I-PER}      & 1.21     & 2.59     & 5.34     & 54.13    & 0.2840     & 0.1376     & 0.2888     \\
\textbf{I-PER+at}   & \two{0.74} & \two{1.46} & 7.81     & 72.85    & 0.2766     & 0.0814     & 0.2860     \\
\hline
\\[-0.2cm]
& \multicolumn{7}{c}{\textbf{MNIST - CNN, ReLU, $l_2$, $\epsilon = 0.3$}} \\[0.1cm]
plain               & 1.28     & 4.93     & 100.00   & 100.00    & 0.0000     & 0.0000     & 0.0000     \\
at                  & 1.12     & 2.50     & 100.00   & 100.00    & 0.0000     & 0.0000     & 0.0000     \\
KW                  & 1.11     & 2.05     & 5.84     & 100.00    & 0.2825     & 0.0000     & 0.2861     \\
IBP                 & 2.37     & 3.85     & 51.12    & \one{11.73} & 0.1534     & \one{0.2648} & 0.1669     \\
C-IBP               & 2.89     & 4.44     & 31.62    & \two{12.29} & 0.2051     & \two{0.2631} & 0.2178     \\
MMR                 & 2.57     & 5.49     & 10.03    & 100.00    & 0.2699     & 0.0000     & 0.2788     \\
MMR+at              & 1.73     & 3.22     & 9.46     & 100.00    & 0.2716     & 0.0000     & 0.2780     \\
\textbf{C-PER}      & 1.02     & 1.87     & \one{5.04} & 100.00    & \one{0.2849} & 0.0000     & \one{0.2882} \\
\textbf{C-PER+at}   & \one{0.43} & \one{0.91} & \two{5.43} & 100.00    & \two{0.2837} & 0.0000     & \two{0.2878} \\
\textbf{I-PER}      & 1.11     & 2.16     & 6.37     & 100.00    & 0.2809     & 0.0000     & 0.2851     \\
\textbf{I-PER+at}   & \two{0.52} & \two{1.12} & 7.89     & 100.00    & 0.2763     & 0.0000     & 0.2812     \\
\hline
\\[-0.2cm]
& \multicolumn{7}{c}{\textbf{CIFAR10 - CNN, ReLU, $l_2$, $\epsilon = 0.1$}} \\[0.1cm]
plain               & 23.29     & 47.39     & 100.00    & 100.00    & 0.0000     & 0.0000     & 0.0000     \\
at                  & 25.84     & 35.81     & 99.96     & 100.00    & 0.0000     & 0.0000     & 0.0000     \\
KW                  & 40.24     & 43.87     & 48.98     & 100.00    & 0.0510     & 0.0000     & 0.0533     \\
IBP                 & 57.90     & 60.03     & 64.78     & \one{78.13} & 0.0352     & \one{0.0219} & 0.0366     \\
C-IBP               & 71.21     & 72.51     & 76.23     & \two{80.97} & 0.0238     & \two{0.0190} & 0.0256     \\
MMR                 & 40.93     & 50.57     & 57.07     & 100.00    & 0.0429     & 0.0000     & 0.0480     \\
MMR+at              & 37.78     & 43.98     & 53.33     & 100.00    & 0.0467     & 0.0000     & 0.0502     \\
\textbf{C-PER}      & 34.10     & 52.54     & 63.42     & 100.00    & 0.0369     & 0.0000     & 0.0465     \\
\textbf{C-PER+at}   & \two{25.76} & \two{33.47} & \one{46.74} & 100.00    & \one{0.0533} & 0.0000     & \one{0.0580} \\
\textbf{I-PER}      & 33.94     & 43.06     & 56.80     & 100.00    & 0.0432     & 0.0000     & 0.0484     \\
\textbf{I-PER+at}   & \one{24.85} & \one{31.32} & \two{47.28} & 100.00    & \two{0.0528} & 0.0000     & \two{0.0572} \\
\Xhline{4\arrayrulewidth}
\end{tabular}
\caption{Full results of 11 training schemes and 7 evaluation schemes for ReLU networks under $l_2$ attacks. The best and the second best results among provably robust training methods (plain and at excluded) are bold. In addition, the best results are underlined.} \label{tbl:main_l2}
\end{table*}

\subsubsection{$l_2$ Robustness on non-ReLU Networks} \label{subsubsec:app_nonrelu_l2}

The results of 8 training methods and 7 evaluation metrics on $l_2$ robustness in the case of Non-ReLU networks are provided in Table~\ref{tbl:nonrelu_l2}.
Consider the best certified robust accuracy for each model, we can clearly see that our proposed PER and PER+at achieve the best performance.
Meanwhile, its clean accuracy is also better than the baselines.
In addition, IBP can only give good performance on IBP-trained models.
These observations are the same as the $l_\infty$ cases.

\begin{table*}[h]
\centering
\small
\begin{tabular}{p{2.2cm}p{1.4cm}<{\raggedleft\arraybackslash}p{1.4cm}<{\raggedleft\arraybackslash}p{1.4cm}<{\raggedleft\arraybackslash}p{1.4cm}<{\raggedleft\arraybackslash}p{1.4cm}<{\raggedleft\arraybackslash}p{1.4cm}<{\raggedleft\arraybackslash}p{1.4cm}<{\raggedleft\arraybackslash}}
\Xhline{4\arrayrulewidth}
Methods & CTE (\%) & PGD (\%) & CRE CRO (\%) & CRE IBP (\%) & ACB CRO & ACB IBP & ACB PEC \\
\hline
\\[-0.2cm]
& \multicolumn{7}{c}{\textbf{MNIST - FC1, Sigmoid, $l_2$, $\epsilon = 0.3$}} \\[0.1cm]
plain              & 2.01     & 10.25     & 30.78     & 94.82    & 0.2077     & 0.0155     & 0.2539     \\
at                 & 1.65     & 3.48      & 7.50      & 85.84    & 0.2775     & 0.0422     & 0.2839     \\
IBP                & 1.40     & 3.07      & 6.43      & 9.13     & 0.2807     & 0.2726     & 0.2873     \\
C-IBP              & 1.51     & 3.24      & 6.36      & \one{8.73} & 0.2709     & \one{0.2738} & 0.2872     \\
\textbf{C-PER}     & 1.36     & 2.58      & 6.12      & 73.71    & 0.2816     & 0.0789     & 0.2867     \\
\textbf{C-PER+at}  & \one{0.46} & \one{1.03}  & 5.26      & 68.94    & 0.2842     & 0.0932     & 0.2905     \\
\textbf{I-PER}     & 1.19     & 2.59      & 6.05      & 70.18    & 0.2818     & 0.0895     & 0.2871     \\
\textbf{I-PER+at}  & 0.49     & 1.16      & \one{5.03}  & 65.79    & \one{0.2849} & 0.1026     & \one{0.2907} \\
\hline
\\[-0.2cm]
& \multicolumn{7}{c}{\textbf{MNIST - FC1, Tanh, $l_2$, $\epsilon = 0.3$}} \\[0.1cm]
plain              & 1.94     & 16.46     & 61.66     & 99.64     & 0.1150     & 0.0011     & 0.1789     \\
at                 & 1.36     & 3.02      & 12.35     & 97.66     & 0.2630     & 0.0070     & 0.2735     \\
IBP                & 1.57     & 3.17      & 7.21      & 10.44     & 0.2784     & 0.2688     & 0.2851     \\
C-IBP              & 1.50     & 3.14      & 6.64      & \one{9.53}  & 0.2801     & \one{0.2714} & 0.2861     \\
\textbf{C-PER}     & 1.31     & 2.47      & \one{5.53}  & 55.17     & \one{0.2834} & 0.1345     & 0.2880     \\
\textbf{C-PER+at}  & 0.58     & 1.30      & 5.89      & 54.88     & 0.2823     & 0.1354     & 0.2885     \\
\textbf{I-PER}     & 1.38     & 2.85      & 5.90      & 45.31     & 0.2823     & 0.1641     & 0.2874     \\
\textbf{I-PER+at}  & \one{0.55} & \one{1.17}  & 5.57      & 53.73     & 0.2833     & 0.1388     & \one{0.2890} \\
\Xhline{4\arrayrulewidth}
\end{tabular}
\caption{Full results of 8 training schemes and 7 evaluation schemes for sigmoid and tanh networks under $l_2$ attacks. The best results among provably robust training methods (plain and at excluded) are bold and underlined.} \label{tbl:nonrelu_l2}
\end{table*}

\subsubsection{Parameter Value Distribution} \label{subsec:app_param_value}

\begin{figure}
\centering
\includegraphics[scale = 0.45]{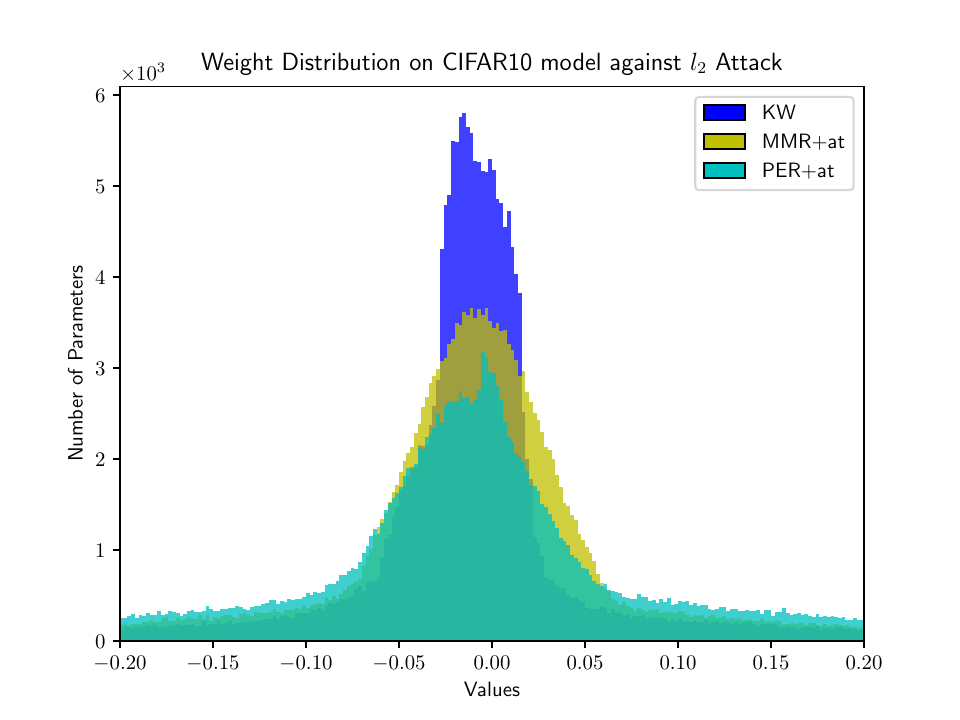}
\caption{Parameter value distributions of CIFAR10 models trained against $l_2$ attack. The Euclidean norms of KW, MMR+at, PER+at model against $l_2$ attacks are 71.34, 62.97 and 141.77, respectively.} \label{fig:param_dist_app}
\end{figure}

The parameter value distributions of CIFAR10 models against $l_2$ attacks are provided in Figure~\ref{fig:param_dist_app}.
Same as the $l_\infty$ cases, the parameters of the PER+at model have significantly larger norms, indicating it better utilize the model's capacity.

\subsubsection{Distribution of the Optimal Bounds} \label{subsec:app_opt_bounds}

\begin{figure}
\centering
\includegraphics[scale = 0.45]{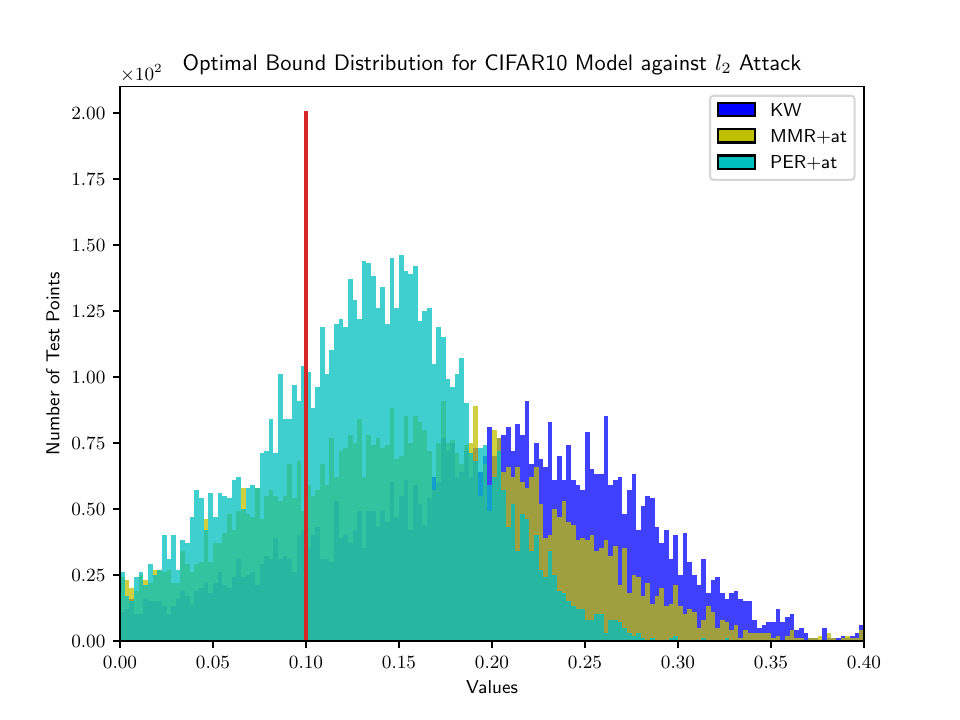}
\caption{Distribution of optimal certified bounds of CIFAR10 models trained against $l_2$ attacks. The target bound ($0.1$) is marked as a red vertical line.}\label{fig:optimal_eps_dist_app}
\end{figure}

The distribution of optimal certified bounds of CIFAR10 models against $l_2$ attacks is shown in Figure~\ref{fig:optimal_eps_dist_app}.
Compared with KW and MMR+at, the values of the optimal certified bounds of the PER+at model are more concentrated in a region slightly better than the required bounds (indicated by the red vertical line).
On the contrary, the KW model usually has unnecessarily large certified bounds on some input instances, indicating over-regularization.

\subsubsection{Searching for the Optimal Value of $\epsilon$} \label{subsec:app_optimal_eps}

Table~\ref{tbl:app_search_1} shows the number of bound calculations in the binary search for the optimal $\epsilon$ in PEC and Fast-Lin under $l_2$ attacks.
The original interval $[\underline{\epsilon}, \bar{\epsilon}]$ is $[0, 1.2]$ for MNIST and $[0, 0.4]$ for CIFAR10.
The bound of the number calculation does not depend on the model in Fast-Lin and is model-dependent in PEC as discussed in Section~\ref{subsec:optimal_budget}.

\begin{table*}[h]
\centering
\small
\begin{tabular}{p{1.4cm}:p{1cm}<{\centering}p{1.15cm}<{\raggedleft\arraybackslash}p{1.15cm}<{\centering}:p{1cm}<{\centering}p{1.15cm}<{\raggedleft\arraybackslash}p{1.15cm}<{\centering}:p{1cm}<{\centering}p{1.15cm}<{\raggedleft\arraybackslash}p{1.15cm}<{\centering}}
\Xhline{4\arrayrulewidth}
\multirow{3}{*}{Methods} & \multicolumn{3}{c}{} & \multicolumn{3}{:c}{} & \multicolumn{3}{:c}{} \\
& \multicolumn{3}{c}{\textbf{MNIST-FC1, ReLU, $l_2$}} & \multicolumn{3}{:c}{\textbf{MNIST-CNN, ReLU, $l_2$}} & \multicolumn{3}{:c}{\textbf{CIFAR10-CNN, ReLU, $l_2$}} \\[0.1cm]
& {\normalsize T\textsubscript{Lin}} & {\normalsize T\textsubscript{PEC}} & {\normalsize $\frac{\text{T}_{\text{PEC}}}{\text{T}_{\text{Lin}}}$} & {\normalsize T\textsubscript{Lin}} & {\normalsize T\textsubscript{PEC}} & {\normalsize $\frac{\text{T}_{\text{PEC}}}{\text{T}_{\text{Lin}}}$} & T\textsubscript{Lin} & {\normalsize T\textsubscript{PEC}} & {\normalsize $\frac{\text{T}_{\text{PEC}}}{\text{T}_{\text{Lin}}}$} \\[0.2cm]
\hline
plain    & \multirow{7}{*}{14} & 9.68  & 0.6914 & \multirow{7}{*}{14} & 13.64 & 0.9742 & \multirow{7}{*}{12} & 11.73 & 0.9775 \\
at       &                     & 10.44 & 0.7457 &                     & 13.76 & 0.9829 &                     & 11.67 & 0.9725 \\
KW       &                     & 7.72  & 0.5514 &                     & 12.63 & 0.9021 &                     & 10.23 & 0.8525 \\
MMR      &                     & 5.86  & 0.4186 &                     & 8.52  & 0.6086 &                     & 9.05  & 0.7542 \\
MMR+at   &                     & 5.91  & 0.4221 &                     & 12.13 & 0.8664 &                     & 10.33 & 0.8608 \\
C-PER    &                     & 11.47 & 0.8194 &                     & 13.75 & 0.9819 &                     & 9.13  & 0.7609 \\
C-PER+at &                     & 11.34 & 0.8100 &                     & 13.72 & 0.9796 &                     & 10.71 & 0.8926 \\
\hline
\end{tabular}
\caption{Number of steps of bound calculation for the optimal $\epsilon$ in Fast-Lin (T\textsubscript{Lin}) and PEC (T\textsubscript{PEC}) for ReLU networks under $l_2$ attacks. Note that T\textsubscript{Lin} is a constant for different models given the original interval $[\underline{\epsilon}, \bar{\epsilon}]$.} \label{tbl:app_search_1}
\end{table*}

%
\IEEEpeerreviewmaketitle

\ifCLASSOPTIONcaptionsoff
  \newpage
\fi

\bibliography{main.bib}
\bibliographystyle{plain}

\end{document}